\documentclass{article}

\usepackage[final,nonatbib]{neurips_2021}

\usepackage[numbers]{natbib}
\usepackage[utf8]{inputenc} 
\usepackage[T1]{fontenc}    
\usepackage{hyperref}       
\usepackage{url}            
\usepackage{booktabs}       
\usepackage{amsmath}
\usepackage{amssymb}
\usepackage{amsfonts}       
\usepackage{nicefrac}       
\usepackage{microtype}      
\usepackage{xcolor}         
\usepackage{subcaption}
\usepackage{algorithm}
\usepackage{algorithmic}

\usepackage{tikz}
\usetikzlibrary{calc,trees,positioning,arrows,chains,shapes.geometric,%
    decorations.pathreplacing,decorations.pathmorphing,shapes,%
    matrix,shapes.symbols}

\newcommand{\revisedfinal}[1]{{#1}}
\newcommand{\revisedmay}[1]{{#1}}

\newcommand{\cellsize}{0.5}
\newcommand{\cellsizecm}{\cellsize cm}

\newcommand{\qnet}{question network}
\newcommand{\touch}{\texttt{touch}}
\newcommand{\name}{rGVFs}

\newcommand{\cutsectionup}{\vspace*{-0.05in}}
\newcommand{\cutsectiondown}{\vspace*{-0.04in}}

\newcommand{\cutcaptionup}{\vspace*{-0.in}}
\newcommand{\cutcaptiondown}{\vspace*{-0.in}}

\title{Learning State Representations from \\
Random Deep Action-conditional Predictions}

\author{%
  Zeyu Zheng \\
  University of Michigan\\
  \texttt{zeyu@umich.edu} \\
   \And
   Vivek Veeriah \\
  University of Michigan\\
  \texttt{vveeriah@umich.edu} \\
   \And
   Risto Vuorio \\
   University of Oxford \\
   \texttt{risto.vuorio@cs.ox.ac.uk} \\
   \And
   Richard Lewis \\
   University of Michigan \\
   \texttt{rickl@umich.edu} \\
   \And
   Satinder Singh \\
  University of Michigan\\
  \texttt{baveja@umich.edu} \\
}

\begin{document}

\maketitle

\begin{abstract}
Our main contribution in this work is an empirical finding that random General Value Functions (GVFs), i.e., deep action-conditional predictions---random both in what feature of observations they predict as well as in the sequence of actions the predictions are conditioned upon---form good auxiliary tasks for reinforcement learning (RL) problems. 
In particular, we show that random deep action-conditional predictions when used as auxiliary tasks yield state representations that produce control performance competitive with state-of-the-art hand-crafted auxiliary tasks like value prediction, pixel control, \revisedfinal{and CURL} in both Atari and DeepMind Lab tasks. In another set of experiments we stop the gradients from the RL part of the network to the state representation learning part of the network and show, perhaps surprisingly, that the auxiliary tasks alone are sufficient to learn state representations good enough to outperform an end-to-end trained actor-critic baseline. 
\revisedfinal{We opensourced our code at \url{https://github.com/Hwhitetooth/random_gvfs}.}
\end{abstract}

\cutsectionup
\section{Introduction
\label{sec:intro}}
\cutsectiondown

Providing auxiliary tasks to Deep Reinforcement Learning (Deep RL) agents has become an important class of methods for driving the learning of state representations that accelerate learning on a main task. Existing auxiliary tasks have the property that their \emph{semantics} are fixed and carefully designed by the agent designer. Some notable examples include pixel control, reward prediction, termination prediction, and multi-horizon value prediction (these are reviewed in more detail below). Unlike the prior approaches that require careful design of auxiliary task semantics, we explore here a different approach in which a set of \emph{random action-conditional prediction tasks} are generated through a rich space of general value functions (GVFs) defined by a language of predictions of random features of observations conditioned on a random sequence of actions. 

Our main, and perhaps surprising, contribution in this work is an empirical finding that auxiliary tasks of learning random GVFs---again, random in both predicted features and actions the predictions are conditioned upon--- yield state representations that produce control performance that is competitive with state-of-the-art auxiliary tasks with hand-crafted semantics. 
We demonstrate this competitiveness in Atari games and DeepMind Lab tasks, comparing to multi-horizon value prediction~\citep{fedus2019hyperbolic}, pixel control~\citep{jaderberg2016reinforcement}, \revisedfinal{and CURL~\citep{laskin2020curl}} as our baseline auxiliary tasks. Note that while we present a reasonable approach to generating the semantics of the random GVFs we employ in our experiments, the specifics of our approach is not by itself a contribution (and thus not evaluated against other approaches to producing semantics for random GVFs), \revisedmay{and alternative reasonable approaches for generating random GVFs could do as well.}

Additionally, through empirical analyses on illustrative domains we show the benefits of exploiting the richness of GVFs---their temporal depth and action-conditionality.  We also provide direct evidence that using random GVFs learns useful representations for the main task through
\emph{stop-gradient} experiments in which the state representations are trained \emph{solely} via the random-GVF auxiliary tasks without using the usual RL learning with rewards to influence  representation learning.
We show that, again, surprisingly, these stop-gradient agents outperform the end-to-end-trained actor-critic baseline.

\cutsectionup
\section{Background and Related Work
\label{sec:related-work}}
\cutsectiondown

\textbf{Horde and PSRs.} Auxiliary tasks were formalized and introduced to RL in~\citep{sutton2011horde} through the Horde 
architecture. Horde is an off-policy learning framework for learning knowledge represented as GVFs from an agent's experience. 
Our work is related to Horde in the use a rich subspace of GVF predictions but differs in that our interest is in the effect of learning these auxiliary predictions on the main task via shared state representations rather than to show the knowledge captured in these GVFs. 
Our work is also related to predictive state representations (PSRs)~\citep{littman2001predictive,singh2004predictive}. PSRs use predictions \emph{as} state representations whereas our work learns latent state representations from predictions.
Recently, in the use of deep neural networks in RL as powerful function approximators, various auxiliary tasks have been proposed to improve the latent state representations of Deep RL agents. We review these auxiliary tasks below. Our work belongs to this family of work in that the auxiliary prediction tasks are used to improve the state representations of Deep RL agents.

\textbf{Auxiliary tasks using predefined GVF targets.}
\textit{UNREAL}~\citep{jaderberg2016reinforcement} uses reward prediction and pixel control and achieved a significant performance improvement in DeepMind Lab but only marginal improvement in Atari. 
Termination prediction~\citep{kartal2019terminal} is shown to be an useful auxiliary task in episodic RL settings. 
SimCore DRAW~\citep{gregor2019shaping} learns a generative model of future observations conditioned on action sequences and uses it as an auxiliary task to shape the agent's belief states in partially observable environments. 
Fedus et al.~\citep{fedus2019hyperbolic} found that simply predicting the returns with multiple different discount factors (MHVP) serves as effective auxiliary tasks. MHVP relies on the availability of rewards and thus is different from our work and other unsupervised auxiliary tasks. 

\textbf{Information-theoretic auxiliary tasks.}
Information-theoretic approaches to auxiliary tasks learn representations that are informative about the future trajectory of these representations as the agent interacts with the environment.  CPC~\citep{van2018representation}, CPC$|$action~\citep{guo2018neural}, ST-DIM~\citep{anand2019unsupervised}, DRIML~\citep{mazoure2020deep}, and \revisedfinal{ATC}~\citep{stooke2021decoupling} apply different forms of temporal contrastive losses to learn predictions in a latent space. 
\revisedfinal{CURL~\citep{laskin2020curl} ignores the long-term future and applies a contrastive loss on the stack of consecutive frames to learn good visual representations. }
PBL~\citep{guo2020bootstrap} focuses on partially observable environments and introduces a separate target encoder to set the prediction targets. The target encoder is trained to distill the learned state representations.  
SPR~\citep{schwarzer2020dataefficient} replaces the target encoder in PBL with a moving average of the state representation function. 
In addition to being predictive, PI-SAC~\citep{lee2020predictive} also enforces the state representations to be compressed. 
SPR and PI-SAC focuses on data efficiency and only conducted experiments under low data budgets.
In these information-theoretic approaches, 
the targets are not GVFs and despite some empirical success, none could learn long-term predictions effectively. This is in contrast to GVF-like predictions which can be effectively learned via TD as in our work as well as the work presented above. 

\textbf{Theory.} A few recent works~\cite{bellemare2019geometric,dabney2020value,lyle2021effect} have studied the optimal representation in RL from a geometric perspective and provided theoretical insights into why predicting GVF-like targets is helpful in learning state representations. Our work is consistent with this theoretical motivation.

\textbf{GVF discovery.} Veeriah et al.~\citep{veeriah2019discovery} used metagradients to discover  simple GVFs (discounted sums of features of observations)
In this work, we 
show that random choices of features and random but rich GVFs are competitive with the state-of-the art of hand-crafted GVFs as auxiliary tasks. 

\revisedmay{
\textbf{GVF RNNs.} 
Rather than using GVFs as auxiliary tasks, General Value Function Networks (GVFNs)~\citep{schlegel2021general} are a new family of recurrent neural networks (RNNs) where each dimension of the hidden state is a GVF prediction. GVFNs are trained by TD instead of truncated backprop through time. Our work relates to GVFNs in that both works use GVFs to shape state representations. However, unlike GVFNs, our work uses GVFs as auxiliary tasks and does not enforce any semantics to the state representations. Moreover, our empirical study mainly focuses on the control setting where the agent needs to maximize its long-term cumulative rewards, whereas~\citep{schlegel2021general} mainly focuses on time series modelling tasks and online prediction tasks and demonstrated the superior performance of GVFNs over conventional RNNs when the truncated input sequences are short during training.
}


\cutsectionup
\section{Method
\label{sec:method}}
\cutsectiondown

\revisedmay{
In this section we first describe the specific GVFs we studied in this work. Then we describe an algorithm for the construction of random GVFs. We finish this section with a description of the agent architecture used in our empirical work.
}

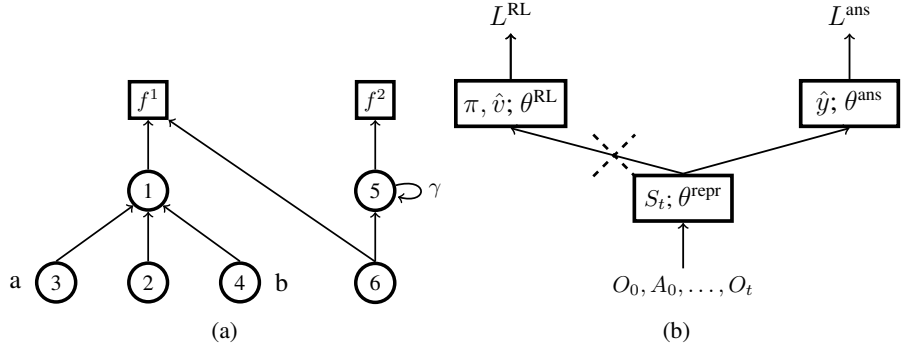
\begin{figure}[tb]
    \centering
    \subfloat[]{\resizebox{0.43\linewidth}{!}{\begin{tikzpicture}[
circlenode/.style={circle, draw=black, ultra thick, minimum size=5mm},
squarednode/.style={rectangle, draw=black, ultra thick, minimum size=5mm},
node distance=0.8cm,
]

\node[squarednode]  (feature_1)         {$f^{1}$};
\node[circlenode] (prediction_1)   [below = of feature_1]   {1};
\node[circlenode] (prediction_2)   [below = of prediction_1]   {2};
\node[circlenode] (prediction_3)   [left=of prediction_2]   {3};
\node[left= 1mm of prediction_3]    {\large a};
\node[circlenode] (prediction_4)   [right=of prediction_2]   {4};
\node[right= 1mm of prediction_4]    {\large b};
\node[squarednode]  (feature_2) [right = 3cm of feature_1]    {$f^{2}$};
\node[circlenode] (prediction_5)   [below = of feature_2]    {5};
\node[circlenode] (prediction_6)   [below = of prediction_5]  {6};

\draw[thick, ->] (prediction_1.north) -- (feature_1.south);
\draw[thick, ->] (prediction_2.north) -- (prediction_1.south);
\draw[thick, ->] (prediction_3.north) -- (prediction_1.south west);
\draw[thick, ->] (prediction_4.north) -- (prediction_1.south east);
\draw[thick, ->] (prediction_5.north) -- (feature_2.south);
\draw[thick, ->] (prediction_5) edge[loop right]node{$\gamma$} (prediction_5);
\draw[thick, ->] (prediction_6.north) -- (prediction_5.south);
\draw[thick, ->] (prediction_6.north) -- (feature_1.south east);

\end{tikzpicture}}\label{fig:question-net-diagram}}
    \subfloat[]{\resizebox{0.43\linewidth}{!}{\begin{tikzpicture}[
dashedcircle/.style={circle, dashed, draw=black, thick, minimum size=5mm},
squarednode/.style={rectangle, draw=black, ultra thick, minimum width=1.5cm, minimum height=7mm},
cross/.style={cross out, draw=black, thick, minimum size=6mm},
node distance=0.7cm and 1cm,
]

\node (history)  {$O_{0}, A_{0}, \dots, O_{t}$};
\node[squarednode]  (representation)    [above=of history]         {$S_{t}$; \large $\theta^{\text{repr}}$};
\node[squarednode]  (rl) [above left=of representation]    {\large $\pi, \hat{v}$; $\theta^{\text{RL}}$};
\node[squarednode]  (auxiliary) [above right=of representation]    {\large $\hat{y}$; $\theta^{\text{ans}}$};
\node (rl_loss) [above=of rl] {\large $L^{\text{RL}}$};
\node (pred_loss) [above =of auxiliary] {\large $L^{\text{ans}}$};

\node[cross, dashed, very thick] (stopgradient) at ($(representation.north)!0.4!(rl.south)$) {};

\draw[thick, ->] (history.north) -- (representation.south);
\draw[thick, ->] (representation.north) -- (rl.south);
\draw[thick, ->] (rl.north) -- (rl_loss.south);
\draw[thick, ->] (rl.north) -- (rl_loss.south);
\draw[thick, ->] (representation.north) -- (auxiliary.south);
\draw[thick, ->] (auxiliary.north) -- (pred_loss.south);

\end{tikzpicture}}\label{fig:agent-arch}}
    \cutcaptionup
    \caption{
    \textbf{(a)} An example of a {\qnet}. The squares represent \emph{feature nodes} and circles represent \emph{prediction nodes}.
    \textbf{(b)} The agent architecture. The dashed cross denotes an optional \texttt{stop-gradient} operation.
    }
    \cutcaptiondown
    
\end{figure}

\subsection{GVFs with Interdependent TD Relationships\label{sec:background_tdnet}}

\revisedmay{
In this work, we study auxiliary prediction tasks where the semantics of the predictions are defined by a set of GVFs with interdependent TD relationships (this family of GVFs are often referred as \emph{temporal-difference networks} in the literature~\citep{sutton2005temporal,sutton2004temporal}). The TD relationships among the GVFs can be described by a graph with directional edges, which we call the \emph{{\qnet}} as it defines the semantics of the predictions.
}

Figure~\ref{fig:question-net-diagram} shows an example of a {\qnet}. 
The two squares represent two \emph{feature nodes} and the six circles represent six \emph{prediction nodes}. 
Node $1$ (labeled in the circles) predicts the expected value of feature $f^{1}$ at the next step. Implicitly, this prediction is conditioned on following the current policy. 
Node $2$ predicts the expected value of node $1$ at the next step. Note that we can ``unroll'' the target of node $2$ to ground it on the features. In this example, node $2$ predicts the expected value of feature $f^{1}$ after two steps when following the current policy. 
Node $5$ has a self-loop and predicts the expectation of the discounted sum of feature $f^{2}$ with a discount factor $\gamma$. We call node $5$ a \emph{discounted sum} prediction node.
Node $3$ is labeled by action $a$. It predicts the expected value of node $1$ at the next step given action $a$ is taken at the current step. We say node $3$ is \emph{conditioned} on action $a$. 
Similarly, node $4$ predicts the same target but is conditioned on action $b$. 
Node $6$ has two outgoing edges. It predicts the \emph{sum} (in general a weighted sum, but in this paper we do not explore the role of these weights and instead fix them to be $1$) of feature $f^{1}$ and the value of node $5$, both at the next step. In this case, it is hard to describe the semantic of node $6$'s prediction in terms of the features, but we can see that the prediction is still grounded on feature $f^{1}$ and $f^{2}$.

Generalising from the example above, a {\qnet} with $n_{p}$ prediction nodes and $n_{f}$ feature nodes defines $n_{p}$ predictions of $n_{f}$ features. We use $N_{p}$ to denote the set of all prediction nodes and $N_{f}$ to denote the set of all feature nodes. Let $W$ be the adjacency matrix of the {\qnet}. $W_{ij}$ denotes the weight on the edge from node $i$ to node $j$. We define $W_{ij} \triangleq 0$ if there is no edge from node $i$ to node $j$. 
Now consider an agent interacting with the environment. At each step $t$, it receives an observation $O_{t}$ and takes an action $A_{t}$ according to its policy $\pi$. Then at the next step it receives an observation $O_{t+1}$. 
The feature $f^{k}(O_{t}, A_{t}, O_{t+1})$ is a scalar function of the transition. The agent makes a prediction $\hat{y}^{i}(O_{0}, A_{0}, \dots, O_{t})$ for each prediction node $i$ based on its history; this is computed by a neural network in our work. 
For brevity, we use $f^{k}_{t+1}$ and $\hat{y}^{i}_{t}$ to denote $f^{k}(O_{t}, A_{t}, O_{t+1})$ and $\hat{y}^{i}(O_{0}, A_{0}, \dots, O_{t})$ respectively.
The target for prediction $i$ at step $t$ is denoted by $y^{i}_{t}$. If prediction node $i$ is not conditioned on any action, its target is
\begin{align*}
    y^{i}_{t} = \mathbb{E}_{\pi} \big[\sum_{j \in N_{p}} W_{ij} y^{j}_{t+1} + \sum_{k \in N_{f}} W_{ik} f^{k}_{t+1} \big]
\end{align*}
otherwise, if it is conditioned on action $a^{i}$, its target is
\begin{align*}
    y^{i}_{t} = \mathbb{E}_{\pi} \big[\sum_{j \in N_{p}} W_{ij} y^{j}_{t+1} + \sum_{k \in N_{f}} W_{ik} f^{k}_{t+1} | A_{t} = a^{i} \big].
\end{align*}
By the construction of the targets, the agent can learn the prediction $\hat{y}^{i}_{t}$ via TD.
If $i$ is not conditioned on any action, then $\hat{y}^{i}_{t}$ is updated by
\begin{align*}
    \hat{y}^{i}_{t} \leftarrow \sum_{j \in N_{p}} W_{ij} \hat{y}^{j}_{t+1} + \sum_{k \in N_{f}} W_{ik} f^{k}_{t+1}
\end{align*}
otherwise, if $i$ is conditioned on action $a^{i}$, then $\hat{y}^{i}_{t}$ is updated by
\begin{align*}
    \hat{y}^{i}_{t} \leftarrow
    \begin{cases}
        \sum\limits_{j \in N_{p}} W_{ij} \hat{y}^{j}_{t+1} + \sum\limits_{k \in N_{f}} W_{ik} f^{k}_{t+1} \hfill \text{ if $A_{t} = a^{i}$} \\
        \hat{y}^{i}_{t} \hfill \text{ otherwise}
    \end{cases}
\end{align*}
In an episodic setting, if the episode terminates at step $T$, we define $y^{i}_{T} \triangleq 0$ and $\hat{y}^{i}_{T} \triangleq 0$ for all prediction nodes $i$.

These GVFs represent a broad class of predictions. Many existing auxiliary prediction tasks can be expressed by a {\qnet}. 
Reward prediction~\citep{jaderberg2016reinforcement} can be represented by a {\qnet} with a single feature node representing the reward and a single prediction node predicting the reward. 
Multi-horizon value prediction~\citep{fedus2019hyperbolic} can be represented by a similar {\qnet} but with multiple self-loop prediction nodes with different discounts. 
Termination prediction~\citep{kartal2019terminal} can be represented by a {\qnet} with a feature node of constant $1$ and a self-loop node with discount $1$.

\subsection{A Random Question Network Generator\label{sec:radar}}


In this work, instead of hand-crafting a new {\qnet} instance as in previous work on the use of predictions for auxiliary tasks, we verify a conjecture that a large number of random deep, action-conditional predictions is enough to drive the learning of good state representations without needing to carefully hand-design the semantics of those predictions. To test this conjecture, we designed a generator of random {\qnet}s from which we can take samples and evaluate their performance as auxiliary tasks. Specifically, we designed a heuristic algorithm that generates {\qnet}s with random \revisedmay{features and random} structures.

\noindent{\bf Random Features.} We use random features, each computed by a scalar function $g^{k}$ with random parameters. For any transition $(O_{t}, A_{t}, O_{t+1})$, the feature is computed as $f^{k}_{t+1} = |g^{k}(O_{t+1}) - g^{k}(O_{t})|$. Instead of directly using the output of $g^{k}$ as the feature, we use the amount of change in $g^{k}$. A similar transformation was used in pixel control~\citep{jaderberg2016reinforcement}.

\noindent{\bf Random Structure.} \revisedmay{
We designed the random question network generator based on the following intuition. 
Each prediction corresponds to first executing an open-loop action sequence then following the agent’s policy. 
Along the trajectory, it accumulates a feature-value (this would be the reward for the standard value function) at each step. 
Depending on the edges in the question network, the accumulated features can be different for different steps. 
As we will illustrate in Section~\ref{sec:illustration}, these predictions can provide rich training signals for learning good representations. 
Specifically, the
} generator takes $5$ arguments as input: number of features $n_{f}$, the discrete action set $\mathcal{A}$, a discount factor $\gamma$, depth $D$, and repeat $R$. Its output is a {\qnet} that contains $n_{f}$ feature nodes as defined above, and $D+1$ layers, each layer contains $R \times |\mathcal{A}|$ prediction nodes except the first layer which contains $n_{f}$ prediction nodes. 
\revisedmay{
We construct the {\qnet} layer by layer incrementally from layer $0$ to layer $D$. 
}
First, layer $0$ has $n_{f}$ feature nodes and $n_{f}$ prediction nodes; each prediction node has an edge to a distinct feature node with weight $1$ on the edge and has a self-loop with weight $\gamma$. Each prediction node in layer $0$ predicts the discounted sum of its corresponding feature and are not conditioned on actions.
\revisedmay{
Then for each layer $l$ $(1 \leq l \leq D)$, we create $R \times |\mathcal{A}|$ prediction nodes. 
}
Each prediction node is conditioned on one action and there are exactly $R$ nodes that are conditioned on the same action. Each prediction node has two edges, one to a random node in layer $l-1$ and one to a random feature node in layer $0$. Note that prediction nodes in layer $1$ do not necessarily connect to a self-loop prediction node in layer $0$ - they may only connect to a feature node. A constraint for preventing duplicated predictions is included so that any two prediction nodes in layer $l$ that are conditioned on the same action cannot connect to the same prediction node in layer $l-1$. 
\revisedmay{
In our preliminary experiments, we tried adding self-loops to deeper layers and allowing denser connections between nodes. Sometimes those additional loops and dense edges caused instability during training. We leave the study of more sophisticated {\qnet} structures to future work. The Appendix includes pseudocode for the random generator algorithm.
}


\subsection{Agent Architecture\label{sec:agent_arch}}
\revisedmay{We used a standard auxiliary-task-augmented agent architecture, as shown in Figure~\ref{fig:agent-arch}}. 
We base our agent on the actor-critic architecture and it consists of $3$ modules. 
The state representation module, parameterized by $\theta^{\text{repr}}$, maps the history of observations and actions $(O_{0}, A_{0}, \dots, O_{t})$ to a state vector $S_{t}$. 
The RL module, parameterized by $\theta^{\text{RL}}$, maps the state vector $S_{t}$ to a policy distribution over the available actions $\pi(\cdot | S_{t})$ and an approximated value function $\hat{v}(S_{t})$. 
The answer network module, parameterized by $\theta^{\text{ans}}$, maps the state vector $S_{t}$ to a set of predictions $\hat{y}(S_{t})$ equal in size to the number of prediction nodes in the {\qnet}. 
\revisedmay{Like in previous work, the augmented agent has more parameters than the base A2C agent due to the answer network module. However, the policy space and the value function space remain the same and these auxiliary parameters are only used for providing richer training signals for the state representation module.}

We trained the network in two separate ways. In the auxiliary task setting, the RL loss $\mathcal{L}^{\text{RL}}$ is backpropagated to update the parameters of the state representation ($\theta^{\text{repr}}$) and the RL ($\theta^{\text{RL}}$) modules, while the answer network loss $\mathcal{L}^{\text{ans}}$ is backpropagated to update the parameters of the answer network ($\theta^{\text{ans}}$) and state representation ($\theta^{\text{repr}}$) modules. Note that the answer network loss only affects the RL module indirectly through the shared state representation module. In the stop-gradient setting, we stopped the gradients from the RL loss from flowing from $\mathcal{L}^{\text{RL}}$ to $\theta^{\text{repr}}$. This allows us to do a harsher and more direct evaluation of how well the auxiliary tasks can train the state representation without any help from the main task. 
For $\mathcal{L}^{\text{RL}}$, we used the standard actor-critic objective with an entropy regularizer. For $\mathcal{L}^{\text{ans}}$, we used the mean-squared loss for all the targets and predictions.

\cutsectionup
\section{Illustrating the Benefits of Deep Action-conditional Questions}
\label{sec:illustration}
\cutsectiondown


\revisedmay{
The main aim of the experiments in this section is to illustrate how deep action-conditional predictions can yield good state representations. We first use a simple grid world to visualize the impact of depth and action conditionality on the learned state representations. Then we demonstrate the practical benefit of exploiting both of these two factors by an ablation study on six Atari games. In addition, to test the robustness of the control performance to the hyperparameters of the random GVFs, we conducted a random search experiment on the Atari game Breakout.
}

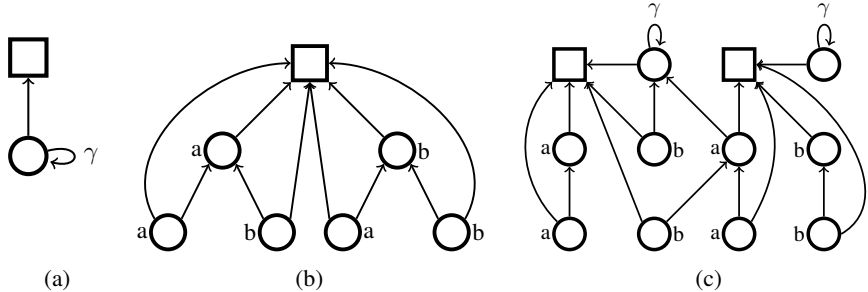
\begin{figure}[tb]
    \centering
    \subfloat[]{\raisebox{1cm}{\resizebox{0.1\linewidth}{!}{\begin{tikzpicture}[
circlenode/.style={circle, draw=black, ultra thick, minimum size=5mm},
squarednode/.style={rectangle, draw=black, ultra thick, minimum size=5mm},
node distance=0.8cm,
]

\node[squarednode]  (feature) {};
\node[circlenode] (td)   [below = of feature]   {};

\draw[thick, ->] (td.north) -- (feature.south);
\draw[thick, ->] (td) edge[loop right]node{$\gamma$} (td);

\end{tikzpicture}}}\label{fig:td-diagram}}
    \subfloat[]{\resizebox{0.38\linewidth}{!}{\begin{tikzpicture}[
circlenode/.style={circle, draw=black, ultra thick, minimum size=5mm},
squarednode/.style={rectangle, draw=black, ultra thick, minimum size=5mm},
node distance=0.8cm,
]

\node[squarednode]  (feature)         {};

\node[circlenode] (depth_1_action_l)   [below left = 0.8cm and 0.8cm of feature]   {};
\node [left=-1mm of depth_1_action_l] {a};

\node[circlenode] (depth_2_action_ll)   [below left = 0.8cm and 0.4cm of depth_1_action_l]   {};
\node [left=-1mm of depth_2_action_ll] {a};
\node[circlenode] (depth_2_action_lr)   [below right = 0.8cm and 0.4cm of depth_1_action_l]   {};
\node [left=-1mm of depth_2_action_lr] {b};

\node[circlenode] (depth_1_action_r)   [below right = 0.8cm and 0.8cm of feature]   {};
\node [right=-1mm of depth_1_action_r] {b};

\node[circlenode] (depth_2_action_rl)   [below left = 0.8cm and 0.4cm of depth_1_action_r]   {};
\node [right=-1mm of depth_2_action_rl] {a};
\node[circlenode] (depth_2_action_rr)   [below right = 0.8cm and 0.4cm of depth_1_action_r]   {};
\node [right=-1mm of depth_2_action_rr] {b};

\draw[thick, ->] (depth_1_action_l.north east) -- (feature.south west);
\draw[thick, ->] (depth_2_action_ll.north east) -- (depth_1_action_l.south west);
\draw[thick, ->] (depth_2_action_ll.north west) to [out=120, in=180] (feature.west);
\draw[thick, ->] (depth_2_action_lr.north west) -- (depth_1_action_l.south east);
\draw[thick, ->] (depth_2_action_lr.north east) -- (feature.south);
\draw[thick, ->] (depth_1_action_r.north west) -- (feature.south east);
\draw[thick, ->] (depth_2_action_rl.north east) -- (depth_1_action_r.south west);
\draw[thick, ->] (depth_2_action_rl.north west) -- (feature.south);
\draw[thick, ->] (depth_2_action_rr.north west) -- (depth_1_action_r.south east);
\draw[thick, ->] (depth_2_action_rr.north east) to [out=60, in=0] (feature.east);

\end{tikzpicture}}\label{fig:tree-diagram}}
    \subfloat[]{\resizebox{0.38\linewidth}{!}{\begin{tikzpicture}[
circlenode/.style={circle, draw=black, ultra thick, minimum size=5mm},
squarednode/.style={rectangle, draw=black, ultra thick, minimum size=5mm},
node distance=0.8cm,
]

\node[squarednode]  (feature_1)         {};

\node[circlenode] (loop_1)   [right = of feature_1]   {};
\draw[thick, ->] (loop_1) edge[loop above]node{$\gamma$} (loop_1);
\draw[thick, ->] (loop_1.west) -- (feature_1.east);

\node[squarednode]  (feature_2) [right = of loop_1]    {};

\node[circlenode] (loop_2)   [right = of feature_2]   {};
\draw[thick, ->] (loop_2) edge[loop above]node{$\gamma$} (loop_2);
\draw[thick, ->] (loop_2.west) -- (feature_2.east);

\node[circlenode] (prediction_1)   [below = of feature_1]   {};
\node [left=-1mm of prediction_1] {a};
\draw[thick, ->] (prediction_1.north) -- (feature_1.south);

\node[circlenode] (prediction_2)   [below = of loop_1]   {};
\node [right=-1mm of prediction_2] {b};
\draw[thick, ->] (prediction_2.north) -- (loop_1.south);
\draw[thick, ->] (prediction_2.north west) -- (feature_1.south east);

\node[circlenode] (prediction_3)   [below = of feature_2]   {};
\node [left=-1mm of prediction_3] {a};
\draw[thick, ->] (prediction_3.north west) -- (loop_1.south east);
\draw[thick, ->] (prediction_3.north) -- (feature_2.south);

\node[circlenode] (prediction_4)   [below = of loop_2]   {};
\node [left=-1mm of prediction_4] {b};
\draw[thick, ->] (prediction_4.north west) -- (feature_2.south east);

\node[circlenode] (prediction_5)   [below = of prediction_1]   {};
\node [left=-1mm of prediction_5] {a};
\draw[thick, ->] (prediction_5.north) -- (prediction_1.south);
\draw[thick, ->] (prediction_5.north west) to [out=135, in=225] (feature_1.south west);

\node[circlenode] (prediction_6)   [below = of prediction_2]   {};
\node [right=-1mm of prediction_6] {b};
\draw[thick, ->] (prediction_6.north east) -- (prediction_3.south west);
\draw[thick, ->] (prediction_6.north west) -- (feature_1.south east);

\node[circlenode] (prediction_7)   [below = of prediction_3]   {};
\node [left=-1mm of prediction_7] {a};
\draw[thick, ->] (prediction_7.north) -- (prediction_3.south);
\draw[thick, ->] (prediction_7.north east) to [out=60, in=300] (feature_2.south east);

\node[circlenode] (prediction_8)   [below = of prediction_4]   {};
\node [left=-1mm of prediction_8] {b};
\draw[thick, ->] (prediction_8.north) -- (prediction_4.south);
\draw[thick, ->] (prediction_8.east) to [out=30, in=345] (feature_2.east);

\end{tikzpicture}}\label{fig:radar-diagram}}
    \cutcaptionup
    \caption{The {\qnet}s we studied in our illustrative experiment. 
    \textbf{(a)} A \revisedmay{discounted sum} prediction. 
    \textbf{(b)} A depth-$2$ tree {\qnet} with $2$ actions. The bottom right prediction node predicts the sum of the values of the feature in the next two steps if action $b$ were taken for both the current and the next step. Other prediction nodes have similar semantics. 
    \textbf{(c)} A random {\qnet} sampled from {\name}. There are $2$ features and $2$ actions, with depth and repeat set to $2$.}
    \cutcaptiondown
    \label{fig:td-net-collection}
\end{figure}

\begin{figure}[tb]
    \centering
    \subfloat[]{\raisebox{0.8cm}{\resizebox{0.18\linewidth}{!}{\input{figures/empty_room}}}\label{fig:empty-room}}
    \subfloat[]{\includegraphics[height=0.28\linewidth]{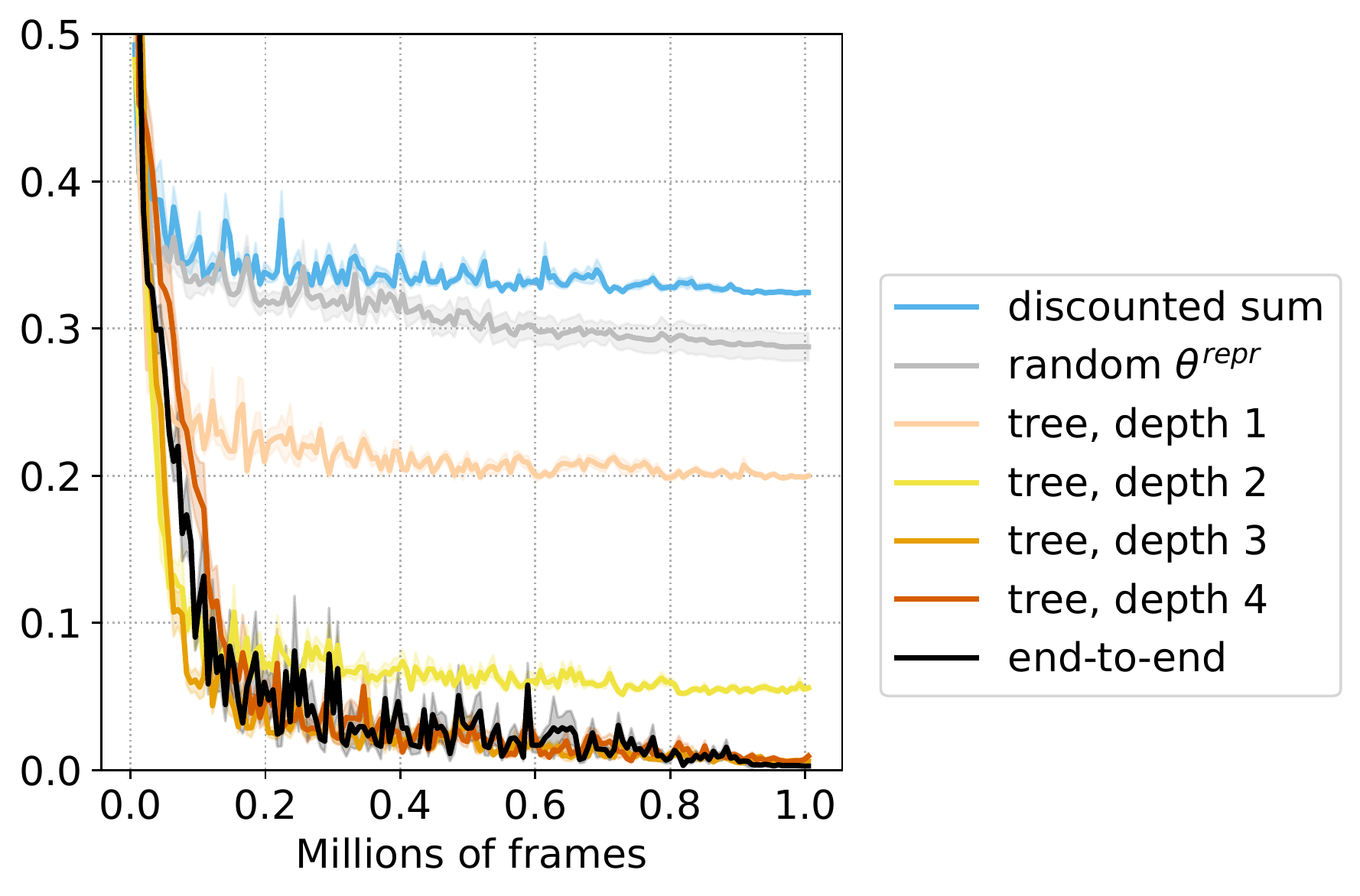}\label{fig:empty-room-touch-loss}}
    \subfloat[]{\includegraphics[height=0.28\linewidth]{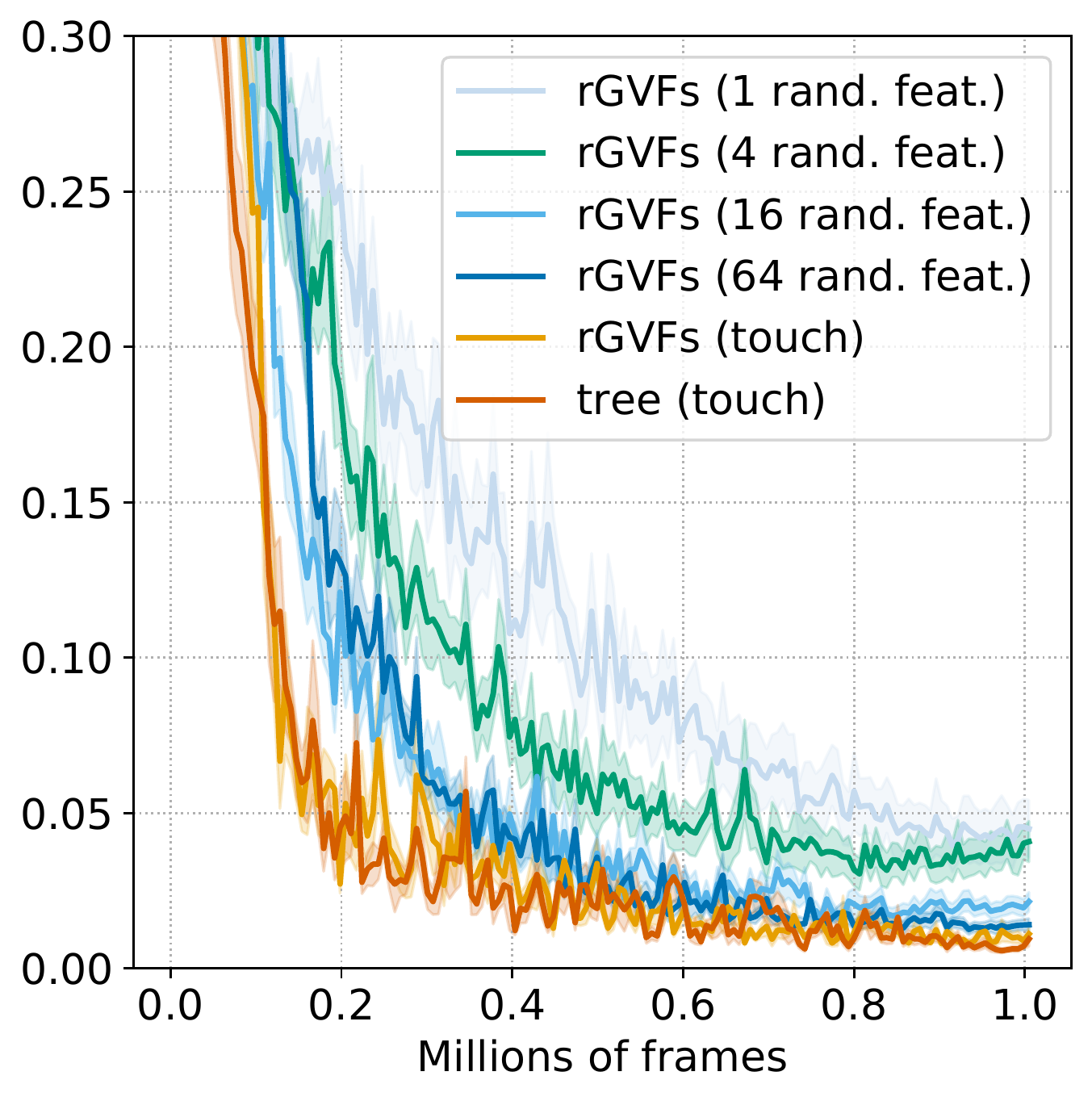}\label{fig:empty-room-rf-rn-loss}}
    \cutcaptionup
    \caption{
    \textbf{(a)} The illustrative grid world environment. The blue circle denotes the agent and the yellow star denotes the rewarding state.
    \textbf{(b)} MSE between the learned value function and the true value function in the tree {\qnet}s experiment. 
    \textbf{(c)} MSE between the learned value function and the true value function in the random {\qnet}s experiment.
    }
    \cutcaptiondown
\end{figure}

\begin{figure}[tb]
    \centering
    \subfloat[]{\includegraphics[height=0.15\linewidth]{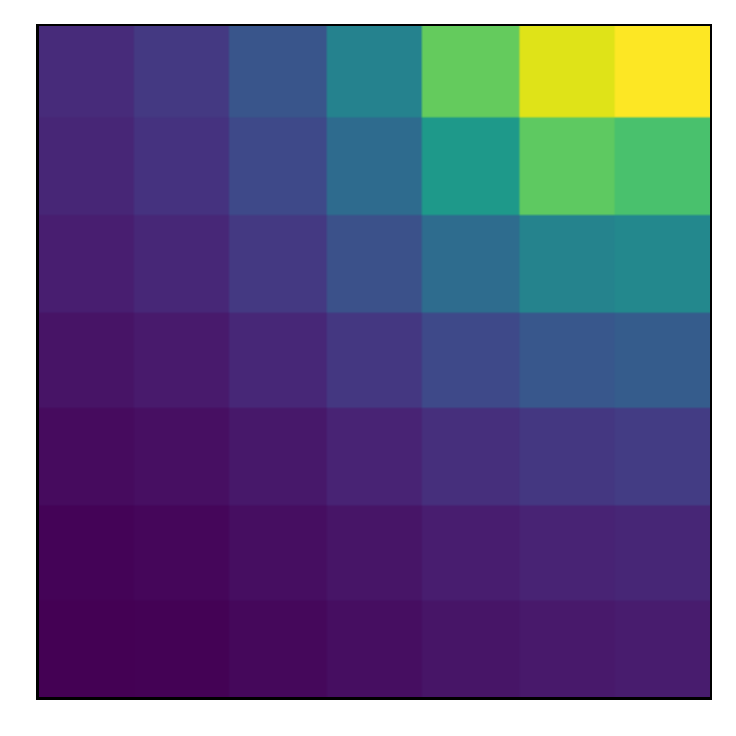}\label{fig:empty-room-truth-heatmap}}
    \subfloat[]{\includegraphics[height=0.15\linewidth]{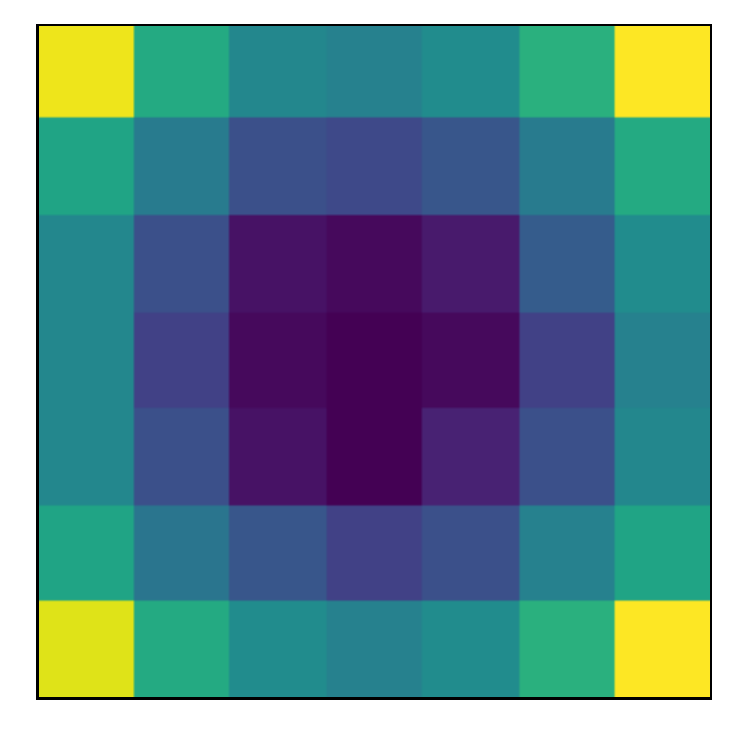}\label{fig:empty-room-td-heatmap}}
    \subfloat[]{\includegraphics[height=0.15\linewidth]{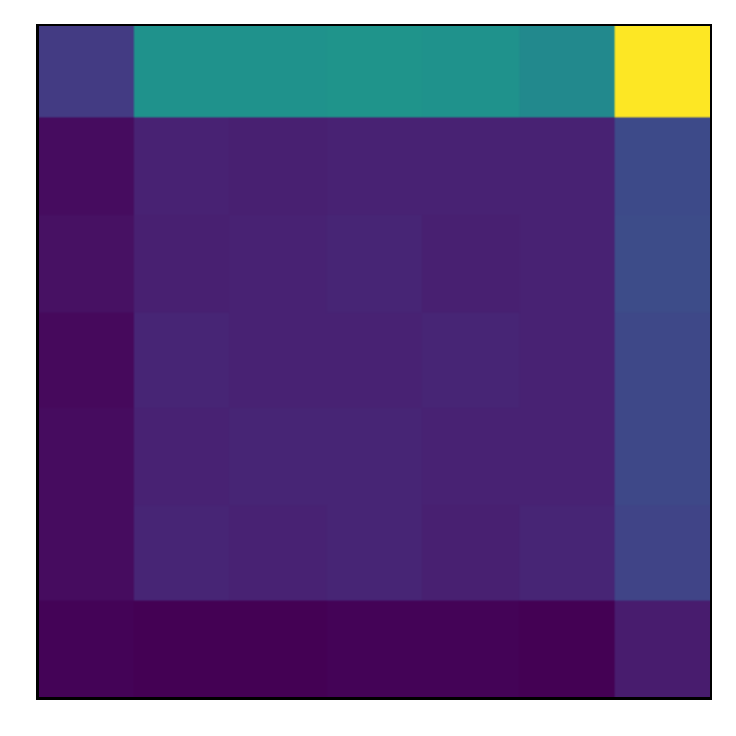}\label{fig:empty-room-tree1-heatmap}}
    \subfloat[]{\includegraphics[height=0.15\linewidth]{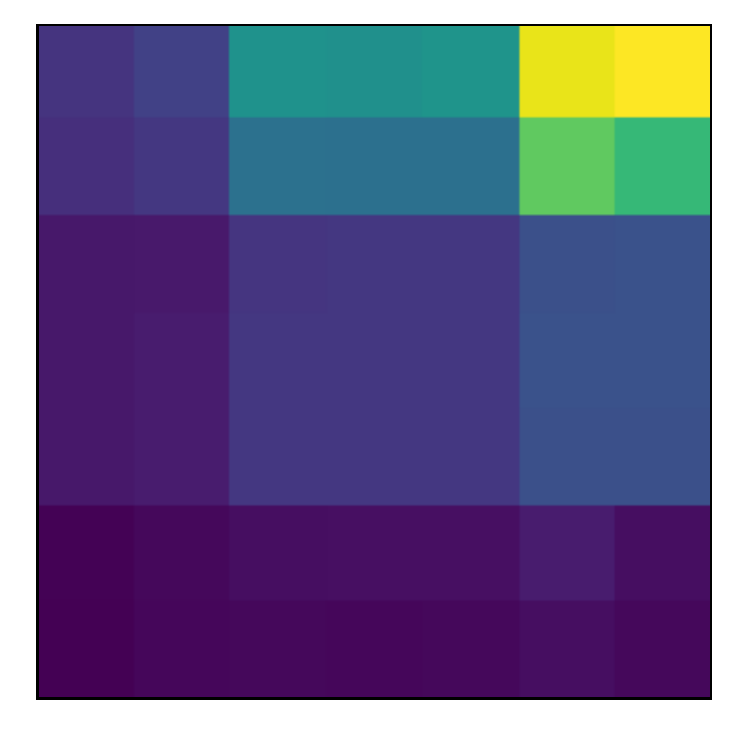}\label{fig:empty-room-tree2-heatmap}}
    \subfloat[]{\includegraphics[height=0.15\linewidth]{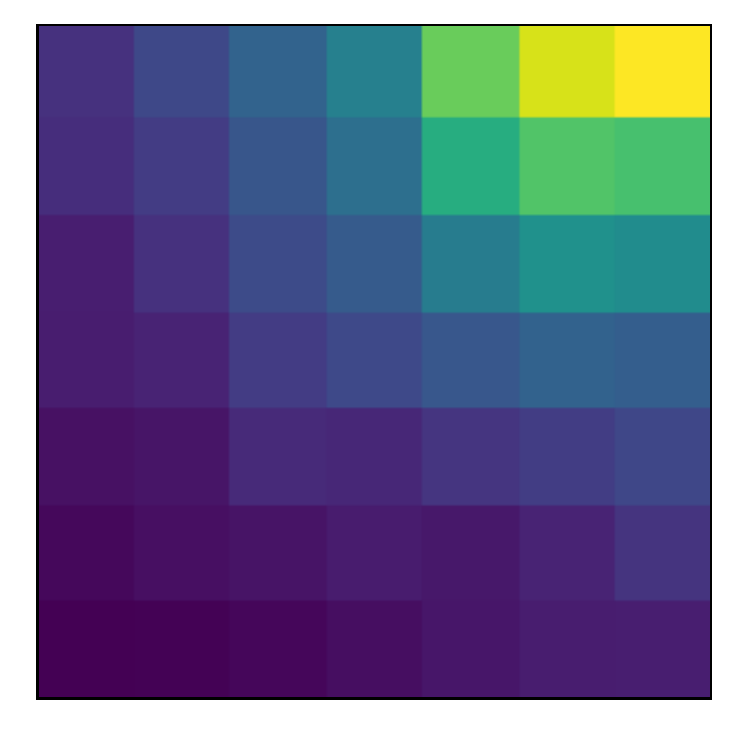}\label{fig:empty-room-tree3-heatmap}}
    \subfloat[]{\includegraphics[height=0.15\linewidth]{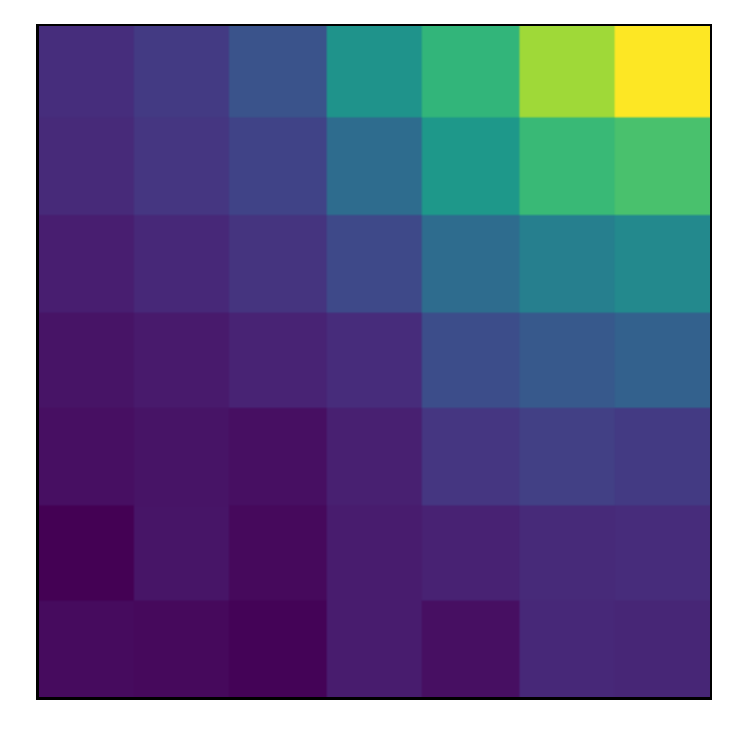}\label{fig:empty-room-tree4-heatmap}}
    \cutcaptionup
    \caption{Visualization of the learned value functions in the empty room environment. Bright indicates high value and dark indicates low value. \textbf{(a)} The true values. 
    \textbf{(b)} The \revisedmay{discounted sum} predictions of the {\touch} feature. \textbf{(c)} - \textbf{(f)} The prediction are defined by a full-tree-structured {\qnet} regarding the {\touch} feature. The depth of the tree increases from $1$ to $4$ from \textbf{(c)} to \textbf{(f)}.}
    \cutcaptiondown
    \label{fig:empty-room-heatmap}
\end{figure}

\subsection{Benefits of Depth \revisedmay{and Action-conditionality}: Illustrative Grid World\label{sec:gridworld_expt_1}}
Although our primary interest (and the focus of subsequent experiments)  is  learning good policies, in this domain we study  \emph{policy evaluation} because this simpler objective is sufficient to illustrate our points and we can compute and visualize the true value function for comparison. Figure~\ref{fig:empty-room} shows the environment,  a $7$ by $7$ grid surrounded by walls. The observation is a top-down view including the walls. 
There are $4$ available actions that move the agent horizontally or vertically to an adjacent cell.
The agent gets a reward of $1$ upon arriving at the goal cell located in the top row, and $0$ otherwise. This is a continuing environment so achieving the goal does not terminate the interaction. 
The objective is to learn the state-value function of a \emph{random policy} which selects each action with equal probability. We used a discount factor of $0.98$.

Specifying a {\qnet} requires specifying both the structure and the features. Later we explore random features, but here we use a single hand-crafted  {\touch} feature so that every prediction has a clear semantics. {\touch} is  $1$ if the agent's move is blocked by the wall and is $0$ otherwise.

Using the {\touch} feature we constructed two types of {\qnet}s. 
The first type is the discounted sum prediction of {\touch} (we used a discount factor $0.8$) (Figure~\ref{fig:td-diagram}). 
The second type is a \emph{full action-conditional tree} of depth $D$. There is only one feature node in the tree which corresponds to the {\touch} feature. Each internal node has $4$ child nodes conditioned on distinct actions. Each prediction node also has a skip edge directly to the feature node (except for the child nodes of the feature node). Figure~\ref{fig:tree-diagram} shows an example of a depth-$2$ tree (the caption describes the semantics of some of the predictions). 
We also compared to a randomly initialized state representation module as a baseline where the state representation was fixed and only the value function weights were learned during training.

\noindent{\bf Neural Network Architecture.}
The empty room environment is fully observable and so the state representation module is a feed-forward neural network that maps the current observation $O_{t}$ to a state vector $S_{t}$. It is parameterized by a $3$-layer multi-layer perceptron (MLP) with $64$ units in the first two layers and $32$ units in the third layer. 
The RL module has one hidden layer with $32$ units and one output head representing the state value. (There is no policy head as the policy was given). 
The answer network module also has one hidden layer with $32$ units and one output layer.
We applied a stop-gradient between the state representation module and the RL module (Figure~\ref{fig:agent-arch}). 
More implementation details are provided in the Appendix.


\noindent{\bf Results.} 
We measured the performance by the mean-squared error (MSE) between the learned value function and the true value function across all states. The true value function was obtained by solving a system of linear equations~\citep{sutton2018reinforcement}.
Figure~\ref{fig:empty-room-touch-loss} shows the MSE during training. Both the random baseline and the discounted sum prediction target performed poorly. 
But even a tree {\qnet} of depth 1 (i.e., four prediction targets corresponding to the four action conditional predictions of {\touch} after one step) performed much better than these two baselines. Performance increased monotonically with increasing depth until depth $3$ when the MSE matched  end-to-end training after $1$ million frames. 

Figure~\ref{fig:empty-room-heatmap} shows the different value functions learned by agents with the different prediction tasks. 
Figure~\ref{fig:empty-room-truth-heatmap} visualizes the true values. 
Figure~\ref{fig:empty-room-td-heatmap} shows the learned value function when the state representations are learned from discounted sum predictions of {\touch}. Its symmetric pattern reflects the symmetry of the grid world and the random policy, but is inconsistent with the asymmetric true values.
Figure~\ref{fig:empty-room-tree1-heatmap} shows the learned value function when the state representations are learned from depth-$1$-tree predictions. It clearly distinguishes $4$ corner states, $4$ groups of states on the boundary, and states in the center area, as this grouping reflects the different prediction targets for these states. 

For the answer network module to make accurate predictions of the targets of the {\qnet}, the state representation module must map states with the same prediction target to similar representations and states with different targets to different representations. As the {\qnet} tree becomes deeper, the agent learns finer state distinctions, until an MSE of $0$ is achieved at depth $3$ (Figure~\ref{fig:empty-room-tree3-heatmap}).


\subsection{Benefits of Random Question Nets: Illustrative Grid World\label{sec:gridworld_expt_2}}
The previous experiment demonstrated benefits of temporally deeper \revisedmay{action-conditonal} prediction tasks. But achieving this by creating deeper and deeper full-branching action-conditional trees is not tractable as the number of prediction targets grows exponentially. The previous experiment also used a single feature formulated using domain knowledge; such feature selection is also not scalable. The \revisedmay{random generator described in Section~\ref{sec:method}} provides a method to mitigate both concerns by growing random {\qnet}s with random features. 

Specifically, we used \emph{discount} $0.8$, \emph{depth} $4$, and \emph{repeat} equal to the number of features \revisedmay{for generating random GVFs}. 
Figure~\ref{fig:empty-room-rf-rn-loss} shows the MSE of different \revisedmay{random GVF} variants. The performance of \revisedmay{random GVFs} with {\touch}---that is, random but not necessarily full branching trees of depth $4$---performed as well as {\touch} with a full tree of depth $4$. \revisedmay{Random GVFs} with a single random feature performed suboptimally; a random feature is likely less discriminative than {\touch}. However, as the number of random features increases, the performance improves, and with $64$ random features, \revisedmay{random GVFs} match the final performance of {\touch} with a full depth $4$ tree.

The results on the grid world provide preliminary evidence that \revisedmay{random deep action-conditional GVFs} with many random features can yield good state representations. We next test this conjecture on a set of Atari games, exploring again the benefits of depth and action conditionality.

\subsection{Ablation Study of Benefits of Depth and Action Conditionality: Atari}
\begin{figure}[tb]
    \centering
    \includegraphics[width=\linewidth]{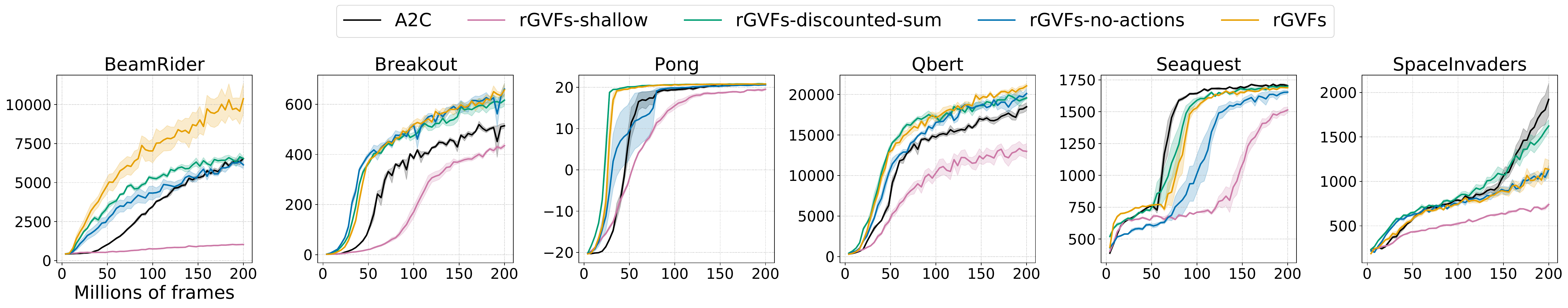}
    \cutcaptionup
    \caption{Learning curves of different {\qnet}s in six Atari games. x-axis denotes the number of frames and y-axis denotes the episode returns. Each curve is averaged over $5$ independent runs with different random seeds. Shaded area shows the standard error.}
    \cutcaptiondown
    \label{fig:atari6-raw-score}
\end{figure}

Here we use six Atari games~\citep{bellemare2013arcade} (these six are often used for hyperparameter selection for the Atari benchmark~\citep{mnih2016asynchronous}) to compare four different kinds of random GVF {\qnet}s: \textbf{(a)} random GVFs in which all predictions are \emph{discounted sums} of distinct random features (illustrated  
 in Figure~\ref{fig:td-diagram} and denoted {\name}-{\it discounted-sum} in Figure~\ref{fig:atari6-raw-score}); \textbf{(b)} random GVFs in which all predictions are \emph{shallow action-conditional} predictions, a set of depth-$1$ trees, each for a distinct random feature (denoted {\name}-{\it shallow} in Figure~\ref{fig:atari6-raw-score}); \textbf{(c)} \revisedmay{random GVFs} without action-conditioning (denoted {\name}-{\it no-actions} in Figure~\ref{fig:atari6-raw-score}); and \textbf{(d)}
random GVFs that exploit both action conditionality and depth (illustrated in Figure~\ref{fig:radar-diagram} and denoted simply by {\name} in Figure~\ref{fig:atari6-raw-score}).

\noindent{\bf Random Features for Atari.}
The random function $g$ for computing the random features are designed as follows. 
The $84 \times 84$ observation $O_{t}$ is divided into $16$ disjoint $21 \times 21$ patches, and a \emph{shared} random linear function applies to each patch to obtain $16$ random features $g^{1}_{t}, g^{2}_{t}, \dots, g^{16}_{t}$. Finally, we process these features as described in \S\ref{sec:radar}.

\noindent{\bf Neural Network Architecture.}
We used A2C~\citep{mnih2016asynchronous} with a standard neural network architecture for Atari~\citep{mnih2015human} as our base agent. 
Specifically, the state representation module consists of $3$ convolutional layers. 
The RL module has one hidden dense layer and two output heads for the policy and the value function respectively. 
The answer network has one hidden dense layer with $512$ units followed by the output layer. 
We stopped the gradient from the RL module to the state representation module.

\noindent{\bf Hyperparameters.}
The discount factor, depth, and repeat were set to $0.95$, $8$, and $16$ respectively. Thus there are $16 + 8 * 16 * |\mathcal{A}|$ total predictions. Random GVFs without action-conditioning has the same {\qnet} except that no prediction was conditioned on actions. To match the total number of predictions, we used $16 + 8 * 16 * |\mathcal{A}|$ random features for discounted sum and $8 * 16$ features for shallow action-conditional predictions. Additional random features were generated by applying more random linear functions to the image patches. The discount factor for discounted sum predictions is also $0.95$. 
More implementation details are provided in the Appendix.

\noindent{\bf Results.} 
Figure~\ref{fig:atari6-raw-score} shows the learning curves in the $6$ Atari games. 
\revisedmay{\name-{\it shallow} performed the worst in all the games, providing further evidence for the value of making deep predictions. 
{\name} consistently outperformed \name-{\it no-actions}, providing  evidence that action-conditioning is beneficial.
And finally, \revisedmay{\name} performed better than \name-{\it discounted-sum} in $3$ out of $6$ games (large difference in BeamRider and small differences in Breakout and Qbert), was comparable in $2$ the other $3$ games, and performed worse in one---despite using many fewer features than \name-{\it discounted-sum}.  This suggests that structured deep action-conditional predictions can be more effective than simply making discounted sum predictions about many features.}

\subsection{Robustness and Stability}
\begin{figure}[tb]
    \centering
    \includegraphics[width=0.85\linewidth]{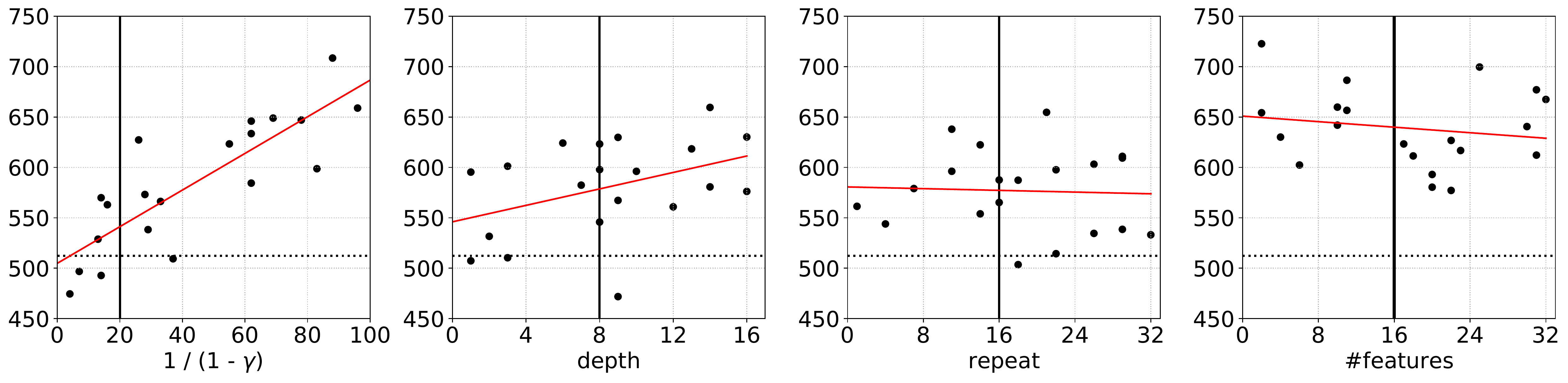}
    \cutcaptionup
    \caption{Scatter plots of scores in Breakout obtained by {\name} with different hyperparameters. x-axis denotes the value of the hyperparameter. y-axis denotes the final game score after training for $200$ million frames. \revisedfinal{The red line in each panel is the line of best fit.} The dotted horizontal lines denote the performance of the end-to-end A2C baseline. The solid vertical lines denotes the values we used in our final experiments.}
    \cutcaptiondown
    \label{fig:breakout-robust}
\end{figure}

We tested the robustness of \revisedmay{\name} with respect to its hyperparameters, namely discount, depth, repeat, and number of features. 
We explored different values for each hyperparameter independently while holding the other hyperparameters fixed to the values we used in the previous experiment. 
For each hyperparameter, we took $20$ samples uniformly from a wide interval and evaluated {\name} on Breakout using the sampled value. 
The results are presented in Figure~\ref{fig:breakout-robust}. 
\revisedfinal{The lines of best fit (the red lines) in the left two panels indicate a positive correlation between the performance and the depth of the predictions, which is consistent with the previous experiments. }
Each hyperparameter has a range of values that achieves high performance, indicating that \revisedmay{{\name} are} stable and robust to different hyperparameter choices. 
\revisedfinal{Additional results in BeamRider and SpaceInvaders are in the Appendix.}

\begin{figure}[tb]
    \centering
    \subfloat[]{\includegraphics[height=0.28\linewidth]{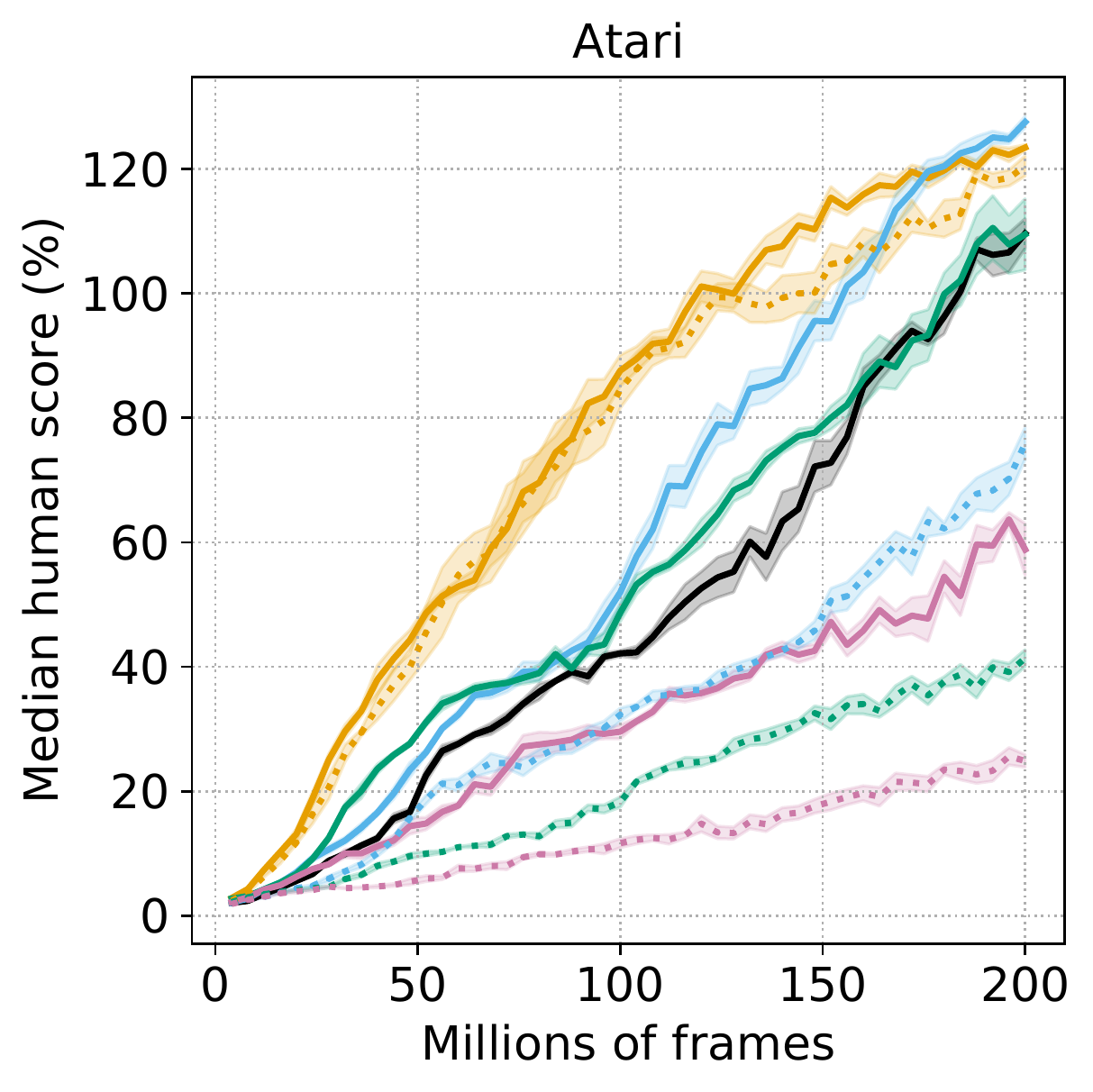}\label{fig:atari-human}}
    \subfloat[]{\includegraphics[height=0.28\linewidth]{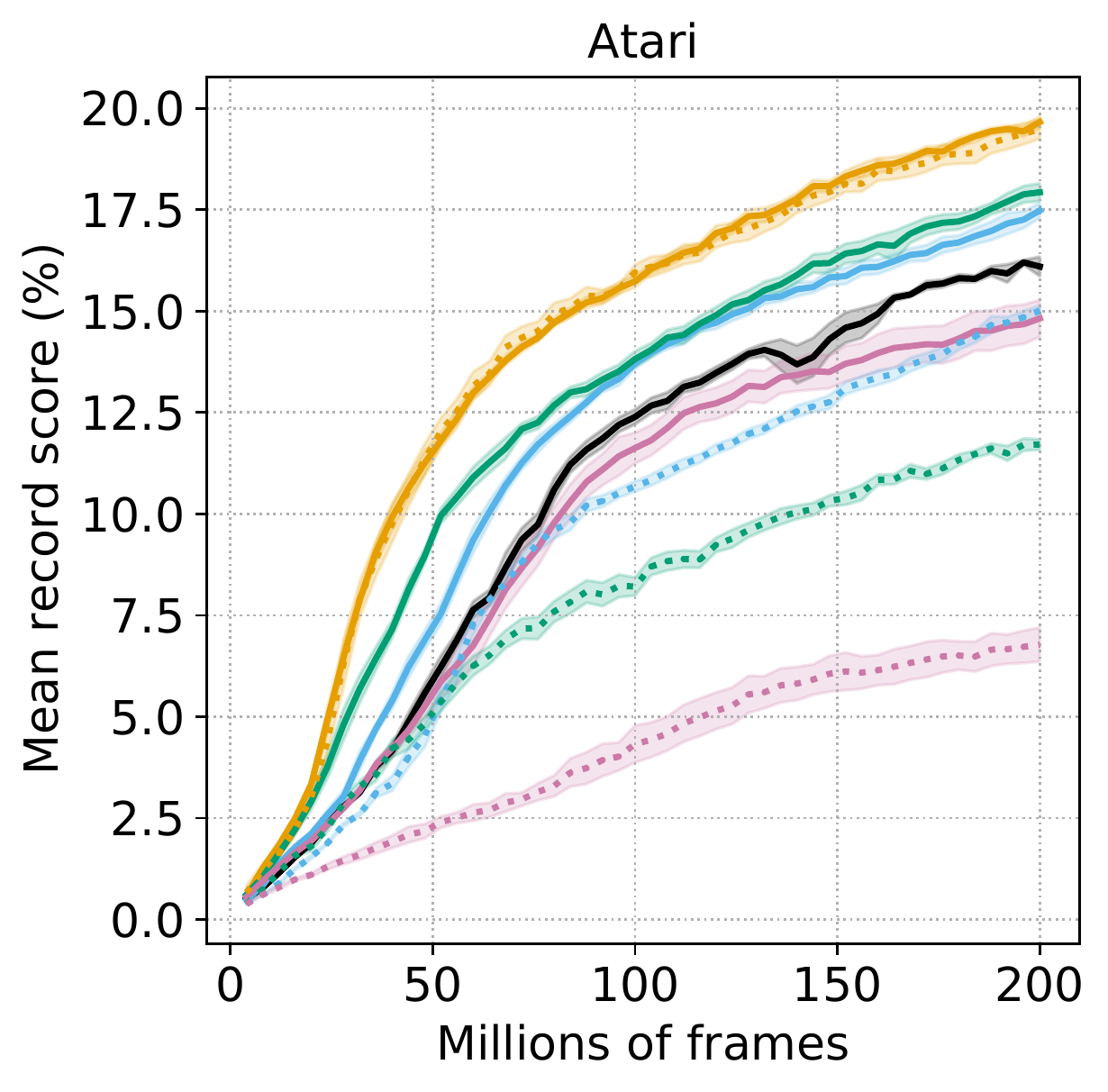}\label{fig:atari-record}}
    \subfloat[]{\includegraphics[height=0.28\linewidth]{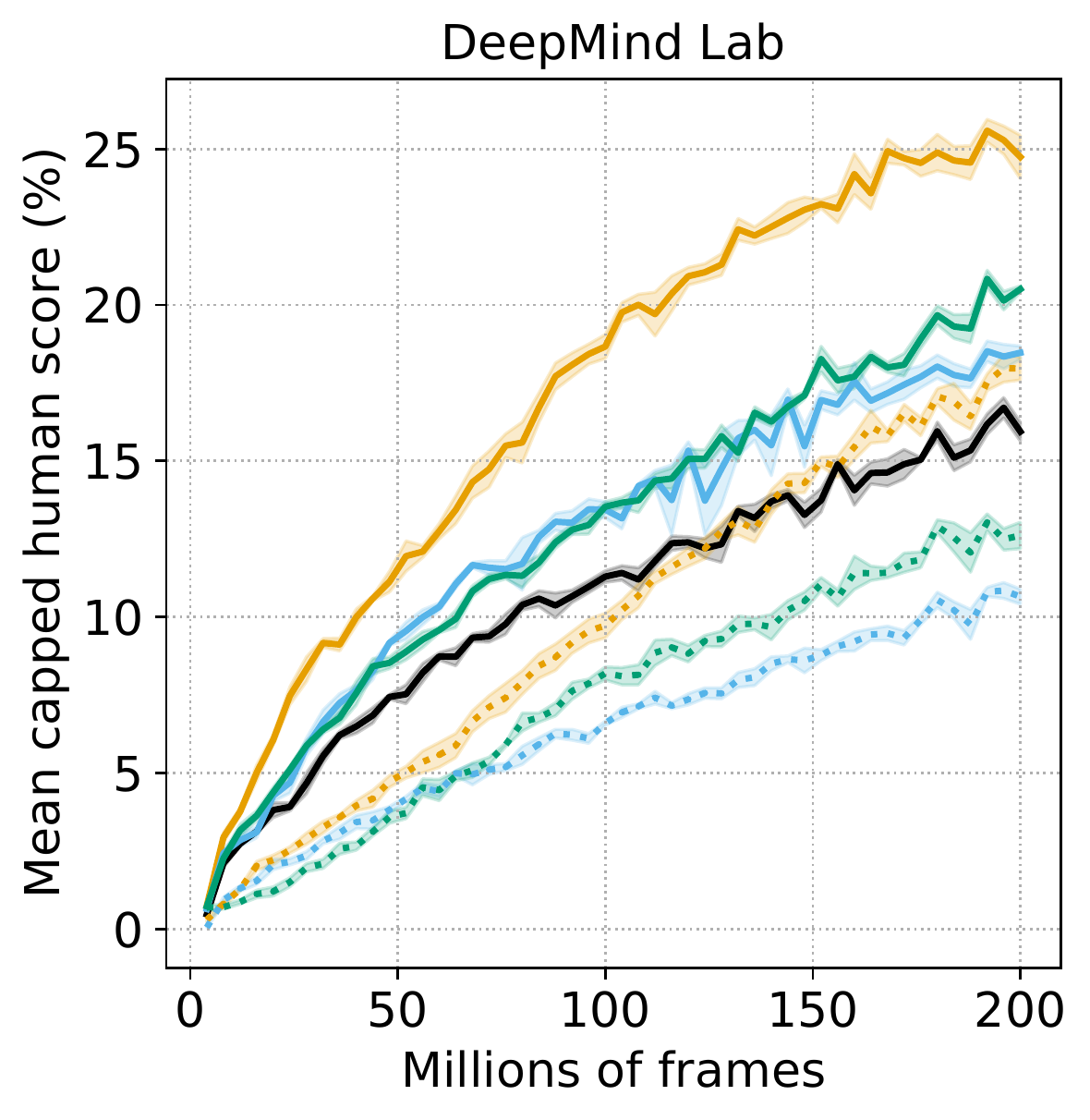}\label{fig:dmlab-human}}
    {\raisebox{0.5\height}{\includegraphics[width=0.13\linewidth]{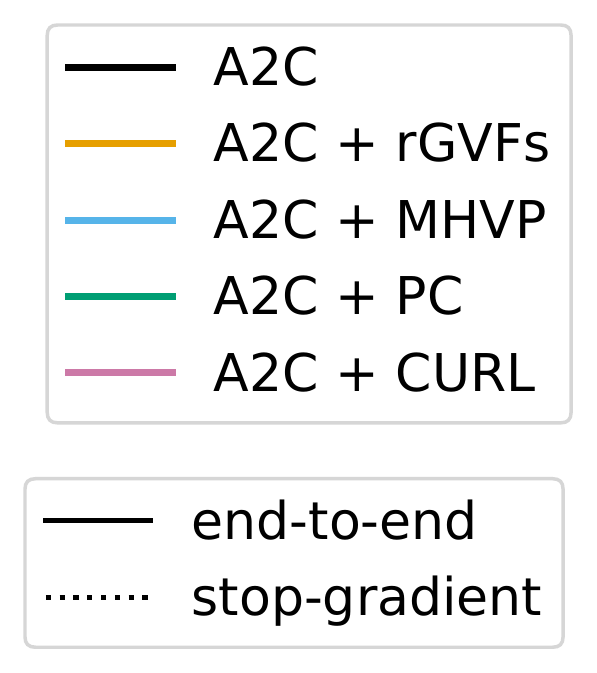}}}
    \cutcaptionup
    \caption{\textbf{(a)} Median human-normalized score across $49$ Atari games. \textbf{(b)} Mean record-normalized score across $49$ Atari games. \textbf{(c)} Mean capped human-normalized score across $12$ DeepMind Lab environments. In all panels, the x-axis denotes the number of frames. Each dark curve is averaged over $5$ independent runs with different random seeds. The shaded area shows the standard error.}
    \cutcaptiondown
    \label{fig:atari-human-score}
\end{figure}

\cutsectionup
\section{Comparison to Baseline Auxiliary Tasks on Atari and DeepMind Lab}
\label{sec:comparison}
\cutsectiondown

In this section, we present the empirical results of comparing the performance of \revisedmay{\name} against the A2C baseline~\citep{mnih2016asynchronous} and \revisedfinal{three other auxiliary tasks, i.e., multi-horizon value prediction (MHVP)~\citep{fedus2019hyperbolic}, pixel control (PC)~\citep{jaderberg2016reinforcement}, and CURL~\citep{laskin2020curl}. }
We conducted the evaluation in $49$ Atari games~\citep{bellemare2013arcade} and $12$ DeepMind Lab environments~\citep{beattie2016deepmind}. 
\revisedfinal{It is unclear how to apply CURL to partially observable environments which require long-term memory because CURL is specifically designed to use the stack of recent frames as the inputs. Thus we did not compare to CURL in the DeepMind Lab environments.
Our implementation of {\name} for this experiment is available at \url{https://github.com/Hwhitetooth/random_gvfs}. }



\noindent{\bf Atari Implementation.}
We used the same architecture for \revisedmay{\name} as in the prior section. 
For MHVP, we used $10$ value predictions following~\citep{fedus2019hyperbolic}. Each prediction has a unique discount factor, chosen to be uniform in terms of their effective horizons from $1$ to $100$ ($\{0, 1 - \frac{1}{10}, 1 - \frac{1}{20}, \dots, 1 - \frac{1}{90}\}$). The architecture for MHVP is the same as \revisedmay{\name}. 
For PC, we followed the architecture design and hyperparameters in ~\citep{jaderberg2016reinforcement}. 
\revisedfinal{For CURL, we implemented it in our experiment setup by using the code accompanying the paper as a reference~\footnote{\url{https://github.com/aravindsrinivas/curl_rainbow}}.}
When not stopping gradient from the RL loss, we mixed the RL updates and the answer network updates by scaling the learning rate for the answer network with a coefficient $c$. We searched $c$ in $\{0.1, 0.2, 0.5, 1, 2\}$ on the $6$ games in the previous section. $c=1$ worked the best for all methods. More details are in the Appendix.

\noindent{\bf DeepMind Lab Implementation.}
We used the same RL module and answer network module as Atari but used a different state representation module to address the partial observability. Specifically, the convolutional layers in the state representation module were followed by a dense layer with $512$ units and a GRU core~\citep{cho2014properties,chung2014empirical} with $512$ units. 

\noindent{\bf Results.}
Figure~\ref{fig:atari-human} and Figure~\ref{fig:atari-record} shows the results for both the stop-gradient and end-to-end architectures on Atari, comparing to two standard human-normalized score measures (median human-normalized score~\citep{mnih2015human} and mean record-normalized score~\citep{hafner2020mastering}).  When training representations end-to-end through a combined main task and auxiliary task loss, the performance of \revisedmay{\name} matches or substantially exceeds the three baselines. 
\revisedfinal{Although the original paper shows that CURL improves agent performance in the data-efficient regime (i.e., $100K$ interactions in Atari), our results indicate that it hurts the performance in the long run. We conjecture that CURL is held back by failing to capture long-term future in representation learning. }
Surprisingly, the stop-gradient \revisedmay{\name} agents outperform the end-to-end A2C baseline, unlike stop-gradient versions of the baseline auxiliary task agents. 
Figure~\ref{fig:dmlab-human} shows the results for both stop-gradient and end-to-end architectures on $12$ DeepMind Lab environments (using mean capped human-normalized scores). Again, \revisedmay{\name} substantially outperforms both auxiliary task baselines, and the stop-gradient version matches the final performance of the end-to-end A2C. Taken together the results from these $61$ tasks provide substantial evidence that \revisedmay{\name} drive the learning of good state representations, outperforming auxiliary tasks with fixed hand-crafted semantics.


\cutsectionup
\section{Conclusion and Future Work}
\label{sec:conclusion}
\cutsectiondown


In this work we provided evidence that learning how to make random deep action-conditional predictions can drive the learning of good state representations. 
\revisedmay{We explored a rich space of GVFs that can be learned efficiently with TD methods.}
Our empirical study on the Atari and DeepMind Lab benchmarks shows that learning state representations solely via auxiliary prediction tasks defined by \revisedmay{random GVFs} outperforms the end-to-end trained A2C baseline. \revisedmay{Random GVFs} also outperformed pixel control, multi-horizon value prediction, \revisedfinal{and CURL} when being used as part of a combined loss function with the  main RL task.

In this work, the {\qnet} was sampled before learning and was held fixed during learning. An interesting goal  for future research is to find methods that can adapt the {\qnet} and discover useful questions  during learning.
The {\qnet}s we studied are limited to discrete actions. It is unclear how to condition a prediction on a continuous action. Thus another future direction to explore is to extend \revisedmay{action-conditional predictions} to continuous action spaces.


\section*{Acknowledgement}
This work was supported by DARPA's L2M program as well as a grant from the Open Philanthropy Project to the Center for Human Compatible AI. 
Any opinions, findings, conclusions, or recommendations expressed here are those of the authors and do not necessarily reflect the views of the sponsors.

\bibliography{references}

\begin{thebibliography}{10}

\bibitem{anand2019unsupervised}
Ankesh Anand, Evan Racah, Sherjil Ozair, Yoshua Bengio, Marc{-}Alexandre
  C{\^{o}}t{\'{e}}, and R.~Devon Hjelm.
\newblock Unsupervised state representation learning in atari.
\newblock In Hanna~M. Wallach, Hugo Larochelle, Alina Beygelzimer, Florence
  d'Alch{\'{e}}{-}Buc, Emily~B. Fox, and Roman Garnett, editors, {\em Advances
  in Neural Information Processing Systems 32: Annual Conference on Neural
  Information Processing Systems 2019, NeurIPS 2019, December 8-14, 2019,
  Vancouver, BC, Canada}, pages 8766--8779, 2019.

\bibitem{beattie2016deepmind}
Charles Beattie, Joel~Z. Leibo, Denis Teplyashin, Tom Ward, Marcus Wainwright,
  Heinrich K{\"{u}}ttler, Andrew Lefrancq, Simon Green, V{\'{\i}}ctor
  Vald{\'{e}}s, Amir Sadik, Julian Schrittwieser, Keith Anderson, Sarah York,
  Max Cant, Adam Cain, Adrian Bolton, Stephen Gaffney, Helen King, Demis
  Hassabis, Shane Legg, and Stig Petersen.
\newblock Deepmind lab.
\newblock {\em CoRR}, abs/1612.03801, 2016.

\bibitem{bellemare2019geometric}
Marc~G. Bellemare, Will Dabney, Robert Dadashi, Adrien~Ali Ta{\"{\i}}ga,
  Pablo~Samuel Castro, Nicolas~Le Roux, Dale Schuurmans, Tor Lattimore, and
  Clare Lyle.
\newblock A geometric perspective on optimal representations for reinforcement
  learning.
\newblock In Hanna~M. Wallach, Hugo Larochelle, Alina Beygelzimer, Florence
  d'Alch{\'{e}}{-}Buc, Emily~B. Fox, and Roman Garnett, editors, {\em Advances
  in Neural Information Processing Systems 32: Annual Conference on Neural
  Information Processing Systems 2019, NeurIPS 2019, December 8-14, 2019,
  Vancouver, BC, Canada}, pages 4360--4371, 2019.

\bibitem{bellemare2013arcade}
Marc~G. Bellemare, Yavar Naddaf, Joel Veness, and Michael Bowling.
\newblock The arcade learning environment: An evaluation platform for general
  agents.
\newblock {\em J. Artif. Intell. Res.}, 47:253--279, 2013.

\bibitem{cho2014properties}
Kyunghyun Cho, Bart van Merrienboer, Dzmitry Bahdanau, and Yoshua Bengio.
\newblock On the properties of neural machine translation: Encoder-decoder
  approaches.
\newblock In Dekai Wu, Marine Carpuat, Xavier Carreras, and Eva~Maria Vecchi,
  editors, {\em Proceedings of SSST@EMNLP 2014, Eighth Workshop on Syntax,
  Semantics and Structure in Statistical Translation, Doha, Qatar, 25 October
  2014}, pages 103--111. Association for Computational Linguistics, 2014.

\bibitem{chung2014empirical}
Junyoung Chung, {\c{C}}aglar G{\"{u}}l{\c{c}}ehre, KyungHyun Cho, and Yoshua
  Bengio.
\newblock Empirical evaluation of gated recurrent neural networks on sequence
  modeling.
\newblock {\em CoRR}, abs/1412.3555, 2014.

\bibitem{dabney2020value}
Will Dabney, Andr{\'{e}} Barreto, Mark Rowland, Robert Dadashi, John Quan,
  Marc~G. Bellemare, and David Silver.
\newblock The value-improvement path: Towards better representations for
  reinforcement learning.
\newblock {\em CoRR}, abs/2006.02243, 2020.

\bibitem{baselines}
Prafulla Dhariwal, Christopher Hesse, Oleg Klimov, Alex Nichol, Matthias
  Plappert, Alec Radford, John Schulman, Szymon Sidor, Yuhuai Wu, and Peter
  Zhokhov.
\newblock Openai baselines.
\newblock \url{https://github.com/openai/baselines}, 2017.

\bibitem{fedus2019hyperbolic}
William Fedus, Carles Gelada, Yoshua Bengio, Marc~G. Bellemare, and Hugo
  Larochelle.
\newblock Hyperbolic discounting and learning over multiple horizons.
\newblock {\em CoRR}, abs/1902.06865, 2019.

\bibitem{gregor2019shaping}
Karol Gregor, Danilo~Jimenez Rezende, Frederic Besse, Yan Wu, Hamza Merzic, and
  A{\"{a}}ron van~den Oord.
\newblock Shaping belief states with generative environment models for {RL}.
\newblock In Hanna~M. Wallach, Hugo Larochelle, Alina Beygelzimer, Florence
  d'Alch{\'{e}}{-}Buc, Emily~B. Fox, and Roman Garnett, editors, {\em Advances
  in Neural Information Processing Systems 32: Annual Conference on Neural
  Information Processing Systems 2019, NeurIPS 2019, December 8-14, 2019,
  Vancouver, BC, Canada}, pages 13475--13487, 2019.

\bibitem{guo2018neural}
Zhaohan~Daniel Guo, Mohammad~Gheshlaghi Azar, Bilal Piot, Bernardo~A. Pires,
  Toby Pohlen, and R{\'{e}}mi Munos.
\newblock Neural predictive belief representations.
\newblock {\em CoRR}, abs/1811.06407, 2018.

\bibitem{guo2020bootstrap}
Zhaohan~Daniel Guo, Bernardo~{\'{A}}vila Pires, Bilal Piot, Jean{-}Bastien
  Grill, Florent Altch{\'{e}}, R{\'{e}}mi Munos, and Mohammad~Gheshlaghi Azar.
\newblock Bootstrap latent-predictive representations for multitask
  reinforcement learning.
\newblock In {\em Proceedings of the 37th International Conference on Machine
  Learning, {ICML} 2020, 13-18 July 2020, Virtual Event}, volume 119 of {\em
  Proceedings of Machine Learning Research}, pages 3875--3886. {PMLR}, 2020.

\bibitem{hafner2020mastering}
Danijar Hafner, Timothy~P. Lillicrap, Mohammad Norouzi, and Jimmy Ba.
\newblock Mastering atari with discrete world models.
\newblock {\em CoRR}, abs/2010.02193, 2020.

\bibitem{jaderberg2016reinforcement}
Max Jaderberg, Volodymyr Mnih, Wojciech~Marian Czarnecki, Tom Schaul, Joel~Z.
  Leibo, David Silver, and Koray Kavukcuoglu.
\newblock Reinforcement learning with unsupervised auxiliary tasks.
\newblock In {\em 5th International Conference on Learning Representations,
  {ICLR} 2017, Toulon, France, April 24-26, 2017, Conference Track
  Proceedings}. OpenReview.net, 2017.

\bibitem{kartal2019terminal}
Bilal Kartal, Pablo Hernandez{-}Leal, and Matthew~E. Taylor.
\newblock Terminal prediction as an auxiliary task for deep reinforcement
  learning.
\newblock In Gillian Smith and Levi Lelis, editors, {\em Proceedings of the
  Fifteenth {AAAI} Conference on Artificial Intelligence and Interactive
  Digital Entertainment, {AIIDE} 2019, October 8-12, 2019, Atlanta, Georgia,
  {USA}}, pages 38--44. {AAAI} Press, 2019.

\bibitem{kingma2015adam}
Diederik~P. Kingma and Jimmy Ba.
\newblock Adam: {A} method for stochastic optimization.
\newblock In Yoshua Bengio and Yann LeCun, editors, {\em 3rd International
  Conference on Learning Representations, {ICLR} 2015, San Diego, CA, USA, May
  7-9, 2015, Conference Track Proceedings}, 2015.

\bibitem{laskin2020curl}
Michael Laskin, Aravind Srinivas, and Pieter Abbeel.
\newblock {CURL:} contrastive unsupervised representations for reinforcement
  learning.
\newblock In {\em Proceedings of the 37th International Conference on Machine
  Learning, {ICML} 2020, 13-18 July 2020, Virtual Event}, volume 119 of {\em
  Proceedings of Machine Learning Research}, pages 5639--5650. {PMLR}, 2020.

\bibitem{lee2020predictive}
Kuang{-}Huei Lee, Ian Fischer, Anthony Liu, Yijie Guo, Honglak Lee, John Canny,
  and Sergio Guadarrama.
\newblock Predictive information accelerates learning in {RL}.
\newblock In Hugo Larochelle, Marc'Aurelio Ranzato, Raia Hadsell,
  Maria{-}Florina Balcan, and Hsuan{-}Tien Lin, editors, {\em Advances in
  Neural Information Processing Systems 33: Annual Conference on Neural
  Information Processing Systems 2020, NeurIPS 2020, December 6-12, 2020,
  virtual}, 2020.

\bibitem{littman2001predictive}
Michael~L. Littman, Richard~S. Sutton, and Satinder Singh.
\newblock Predictive representations of state.
\newblock In Thomas~G. Dietterich, Suzanna Becker, and Zoubin Ghahramani,
  editors, {\em Advances in Neural Information Processing Systems 14 [Neural
  Information Processing Systems: Natural and Synthetic, {NIPS} 2001, December
  3-8, 2001, Vancouver, British Columbia, Canada]}, pages 1555--1561. {MIT}
  Press, 2001.

\bibitem{lyle2021effect}
Clare Lyle, Mark Rowland, Georg Ostrovski, and Will Dabney.
\newblock On the effect of auxiliary tasks on representation dynamics.
\newblock In Arindam Banerjee and Kenji Fukumizu, editors, {\em The 24th
  International Conference on Artificial Intelligence and Statistics, {AISTATS}
  2021, April 13-15, 2021, Virtual Event}, volume 130 of {\em Proceedings of
  Machine Learning Research}, pages 1--9. {PMLR}, 2021.

\bibitem{mazoure2020deep}
Bogdan Mazoure, Remi~Tachet des Combes, Thang Doan, Philip Bachman, and
  R.~Devon Hjelm.
\newblock Deep reinforcement and infomax learning.
\newblock In Hugo Larochelle, Marc'Aurelio Ranzato, Raia Hadsell,
  Maria{-}Florina Balcan, and Hsuan{-}Tien Lin, editors, {\em Advances in
  Neural Information Processing Systems 33: Annual Conference on Neural
  Information Processing Systems 2020, NeurIPS 2020, December 6-12, 2020,
  virtual}, 2020.

\bibitem{mnih2016asynchronous}
Volodymyr Mnih, Adri{\`{a}}~Puigdom{\`{e}}nech Badia, Mehdi Mirza, Alex Graves,
  Timothy~P. Lillicrap, Tim Harley, David Silver, and Koray Kavukcuoglu.
\newblock Asynchronous methods for deep reinforcement learning.
\newblock In Maria{-}Florina Balcan and Kilian~Q. Weinberger, editors, {\em
  Proceedings of the 33nd International Conference on Machine Learning, {ICML}
  2016, New York City, NY, USA, June 19-24, 2016}, volume~48 of {\em {JMLR}
  Workshop and Conference Proceedings}, pages 1928--1937. JMLR.org, 2016.

\bibitem{mnih2015human}
Volodymyr Mnih, Koray Kavukcuoglu, David Silver, Andrei~A. Rusu, Joel Veness,
  Marc~G. Bellemare, Alex Graves, Martin~A. Riedmiller, Andreas Fidjeland,
  Georg Ostrovski, Stig Petersen, Charles Beattie, Amir Sadik, Ioannis
  Antonoglou, Helen King, Dharshan Kumaran, Daan Wierstra, Shane Legg, and
  Demis Hassabis.
\newblock Human-level control through deep reinforcement learning.
\newblock {\em Nat.}, 518(7540):529--533, 2015.

\bibitem{schlegel2021general}
Matthew Schlegel, Andrew Jacobsen, Zaheer Abbas, Andrew Patterson, Adam White,
  and Martha White.
\newblock General value function networks.
\newblock {\em J. Artif. Intell. Res.}, 70:497--543, 2021.

\bibitem{schwarzer2020dataefficient}
Max Schwarzer, Ankesh Anand, Rishab Goel, R~Devon Hjelm, Aaron Courville, and
  Philip Bachman.
\newblock Data-efficient reinforcement learning with self-predictive
  representations.
\newblock {\em arXiv preprint arXiv:2007.05929}, 2020.

\bibitem{singh2004predictive}
Satinder Singh, Michael~R. James, and Matthew~R. Rudary.
\newblock Predictive state representations: {A} new theory for modeling
  dynamical systems.
\newblock In David~Maxwell Chickering and Joseph~Y. Halpern, editors, {\em
  {UAI} '04, Proceedings of the 20th Conference in Uncertainty in Artificial
  Intelligence, Banff, Canada, July 7-11, 2004}, pages 512--518. {AUAI} Press,
  2004.

\bibitem{stooke2021decoupling}
Adam Stooke, Kimin Lee, Pieter Abbeel, and Michael Laskin.
\newblock Decoupling representation learning from reinforcement learning.
\newblock In Marina Meila and Tong Zhang, editors, {\em Proceedings of the 38th
  International Conference on Machine Learning, {ICML} 2021, 18-24 July 2021,
  Virtual Event}, volume 139 of {\em Proceedings of Machine Learning Research},
  pages 9870--9879. {PMLR}, 2021.

\bibitem{sutton2018reinforcement}
Richard~S Sutton and Andrew~G Barto.
\newblock {\em Reinforcement learning: An introduction}.
\newblock MIT press, 2018.

\bibitem{sutton2011horde}
Richard~S. Sutton, Joseph Modayil, Michael Delp, Thomas Degris, Patrick~M.
  Pilarski, Adam White, and Doina Precup.
\newblock Horde: a scalable real-time architecture for learning knowledge from
  unsupervised sensorimotor interaction.
\newblock In Liz Sonenberg, Peter Stone, Kagan Tumer, and Pinar Yolum, editors,
  {\em 10th International Conference on Autonomous Agents and Multiagent
  Systems {(AAMAS} 2011), Taipei, Taiwan, May 2-6, 2011, Volume 1-3}, pages
  761--768. {IFAAMAS}, 2011.

\bibitem{sutton2005temporal}
Richard~S. Sutton, Eddie~J. Rafols, and Anna Koop.
\newblock Temporal abstraction in temporal-difference networks.
\newblock In {\em Advances in Neural Information Processing Systems 18 [Neural
  Information Processing Systems, {NIPS} 2005, December 5-8, 2005, Vancouver,
  British Columbia, Canada]}, pages 1313--1320, 2005.

\bibitem{sutton2004temporal}
Richard~S. Sutton and Brian Tanner.
\newblock Temporal-difference networks.
\newblock In {\em Advances in Neural Information Processing Systems 17 [Neural
  Information Processing Systems, {NIPS} 2004, December 13-18, 2004, Vancouver,
  British Columbia, Canada]}, pages 1377--1384, 2004.

\bibitem{van2018representation}
A{\"{a}}ron van~den Oord, Yazhe Li, and Oriol Vinyals.
\newblock Representation learning with contrastive predictive coding.
\newblock {\em CoRR}, abs/1807.03748, 2018.

\bibitem{veeriah2019discovery}
Vivek Veeriah, Matteo Hessel, Zhongwen Xu, Janarthanan Rajendran, Richard~L.
  Lewis, Junhyuk Oh, Hado van Hasselt, David Silver, and Satinder Singh.
\newblock Discovery of useful questions as auxiliary tasks.
\newblock In Hanna~M. Wallach, Hugo Larochelle, Alina Beygelzimer, Florence
  d'Alch{\'{e}}{-}Buc, Emily~B. Fox, and Roman Garnett, editors, {\em Advances
  in Neural Information Processing Systems 32: Annual Conference on Neural
  Information Processing Systems 2019, NeurIPS 2019, December 8-14, 2019,
  Vancouver, BC, Canada}, pages 9306--9317, 2019.

\bibitem{wang2016dueling}
Ziyu Wang, Tom Schaul, Matteo Hessel, Hado van Hasselt, Marc Lanctot, and Nando
  de~Freitas.
\newblock Dueling network architectures for deep reinforcement learning.
\newblock In Maria{-}Florina Balcan and Kilian~Q. Weinberger, editors, {\em
  Proceedings of the 33nd International Conference on Machine Learning, {ICML}
  2016, New York City, NY, USA, June 19-24, 2016}, volume~48 of {\em {JMLR}
  Workshop and Conference Proceedings}, pages 1995--2003. JMLR.org, 2016.

\end{thebibliography}
\bibliographystyle{plain}
\appendix

\section{Potential Negative Societal Impact}
While all AI advances can have potential negative impact on society through their misuse, this work advances our understanding of fundamental questions of interest to RL and at least at this point is far away from potential misuse.

\section{Implementation Details}

\subsection{Experiments on the Empty Room Environment}
\noindent{\bf Neural Network Architecture.}
The empty room environment is fully observable and so the state representation module is a feed-forward neural network that maps the current observation $O_{t}$ to a state vector $S_{t}$. It is parameterized by a $3$-layer multi-layer perceptron (MLP) with $64$ units in the first two layers and $32$ units in the third layer. 
The RL module has one hidden layer with $32$ units and one output head representing the state value. (There is no policy head as the policy was given). 
The answer network module also has one hidden layer with $32$ units and one output layer. 
ReLU activation is applied after every hidden layer. 
We applied a stop-gradient between the state representation module and the RL module. 

\noindent{\bf Hyperparameters.}
Both the value function and the answer network were updated via TD. We used $8$ parallel actors to generate data and updated the parameters every $8$ steps. We used the Adam optimizer~\citep{kingma2015adam}. 
\revisedmay{We searched the learning rate in $\{0.01, 0.001, 0.0001, 0.00001\}$ and selected $0.001$ for all agents except the end-to-end agent which used $0.0001$.} The value function updates and the answer network updates used two separate optimizers with identical hyperparameters. 

\subsection{Atari Experiments}
\noindent{\bf Neural Network Architecture.}
We used A2C~\citep{mnih2016asynchronous} with a standard neural network architecture for Atari~\citep{mnih2015human} as our base agent. 
Specifically, the state representation module consists of $3$ convolutional layers. The first layer has $32$ $8 \times 8$ convolutional kernels with a stride of $4$, the second layer has $64$ $4 \times 4$ kernels with stride $2$, and the third layer has $64$ $3 \times 3$ kernels with stride $1$. 
The RL module has one dense layer with $512$ units and two output heads for the policy and the value function respectively. 
The answer network has one hidden dense layer with $512$ units followed by the output layer. 
ReLU activation is applied after every hidden layer. 
We stopped the gradient from the RL module to the state representation module.

\noindent{\bf Hyperparameters.}
Following convention~\citep{mnih2015human}, we used a stack of the latest $4$ frames as the input to the agent, i.e., the input to the state representation module at step $t$ is $(O_{t-3}, O_{t-2}, O_{t-1}, O_{t})$. 
We used $16$ parallel actors to generate data and updated the agent's parameters every $20$ steps. 
The entropy regularization was $0.01$ and the discount factor for the A2C loss was $0.99$. We used the RMSProp optimizer with learning rate $0.0007$, decay $0.99$, and $\epsilon=0.00001$. The RL updates and the answer network updates used two separate optimizers with identical hyperparameters. 
\revisedmay{We used two separate optimizers because the gradients from the RL loss and the gradients from the auxiliary loss may have different statistics.} 
The gradient from the A2C loss was clipped by global norm to $0.5$. 
\revisedmay{The values for the above hyperparameters are taken from a well-tuned open-source implementation of A2C for Atari~\citep{baselines}. These values are used for all methods.} 
When not stopping gradient from the RL loss, we mixed the RL updates and the answer network updates by scaling the learning rate for the answer network with a coefficient $c$. We searched $c$ in $\{0.1, 0.2, 0.5, 1, 2\}$ on the $6$ games in the main text. $c=1$ worked the best for both {\name} and baseline methods. 
\revisedmay{The mixing coefficient was applied to the learning rate for the auxiliary loss because we used separate optimizers.}

\noindent{\bf Baseline Methods.}
For MHVP, we used $10$ value predictions following~\citep{fedus2019hyperbolic}. Each prediction has a unique discount factor, chosen to be uniform in terms of their effective horizons from $1$ to $100$ ($\{0, 1 - \frac{1}{10}, 1 - \frac{1}{20}, \dots, 1 - \frac{1}{90}\}$). The architecture for MHVP is the same as {\name}. 
\revisedmay{We tried scaling MHVP up to $1024$ predictions in the six Atari games but did not observe any significant performance improvement. Thus we decided to follow the original work.}
For PC, We followed the architecture design and hyperparameters used in the original work~\citep{jaderberg2016reinforcement}. 
Specifically, we center-cropped the observation image to $80 \times 80$ and used $4 \times 4$ patches which resulted in $400$ features. The discount factor was set to $0.9$. The output of the state representation module is first mapped to a $2592$-dimensional vector by a dense layer and then reshaped to a $32 \times 9 \times 9$ tensor. A deconvolutional layer then maps this $3$D tensor to $|\mathcal{A}| \times 20 \times 20$ representing the action-values for each patch. Following~\citep{jaderberg2016reinforcement}, we used the dueling architecture~\citep{wang2016dueling}. 
\revisedfinal{For CURL, we followed the code accompanying the original paper~\footnote{\url{https://github.com/aravindsrinivas/curl_rainbow}}. We used random crop as the sole data augmentation. The $(84, 84, 4)$ stacked observation was first randomly cropped in to $(80, 80, 4)$ and then zero-padded to the original size. The output layer of the answer network has $128$ units. The bilinear similarity was computed by a $128 \times 128$ matrix whose weights were learned together with the agent parameters. The exponential-moving-average target network used a step size of $0.001$.}

\subsection{DeepMind Lab Experiments}
\noindent{\bf Neural Network Architecture.}
For DeepMind Lab, we used the same RL module and answer network module as Atari but used a different state representation module to address the partial observability. Specifically, the convolutional layers in the state representation module were followed by a dense layer with $512$ units and a GRU core~\citep{cho2014properties,chung2014empirical} with $512$ units. 

\noindent{\bf Hyperparameters.}
We used $32$ parallel actors to generate data and updated the agent's parameters every $20$ steps. The discount factor was $0.99$. 
\revisedmay{We searched the entropy regularization in $\{0.001, 0.003, 0.01, 0.03, 0.1\}$ for the A2C baseline on \emph{explore\_goal\_locations\_small}, \emph{explore\_object\_locations\_small}, and \emph{lasertag\_three\_opponents\_small} and selected $0.003$.} We used the RMSProp optimizer with decay $0.99$ and $\epsilon=10^{-8}$ without tuning. \revisedmay{We searched the learning rate in $\{0.00003, 0.0001, 0.0003, 0.001\}$ for the A2C baseline and selected $0.0003$.} 
The gradient from the A2C loss was clipped by global norm to $0.5$. 
\revisedmay{The values for the above hyperparameters are used for all methods.} 
The RL updates and the answer network updates used two separate optimizers with identical hyperparameters.
\revisedmay{We used two separate optimizers because the gradients from the RL loss and the gradients from the auxiliary loss may have different statistics.} 
For end-to-end (not stop-gradient) agents,  we searched the mixing coefficient $c$ in $\{0.1, 0.2, 0.5, 1, 2\}$ on \emph{explore\_goal\_locations\_small}, \emph{explore\_object\_locations\_small}, and \emph{lasertag\_three\_opponents\_small}. $c=1$ worked the best for both {\name} and baseline methods. 
\revisedmay{The mixing coefficient was applied to the learning rate for the auxiliary loss because we used separate optimizers.}

\subsection{Hardware}
Each agent was trained on a single NVIDIA GeForce RTX $2080$ Ti in all of our experiments.

\revisedfinal{
\subsection{Computational Cost}
Like all auxiliary task methods, the additional computation of the GVF predictions introduces an overhead to the overall training pipeline. 
In our experiment, we found that the overhead was tiny. {\name} was only $6\%$ slower than the A2C baseline because the bottleneck was the agent-environment interaction rather than the parameter updates.
}

\clearpage
\section{Pseudocode for The Random Question Network Generator}

\begin{algorithm}[h!]
\caption{A Random Question Network Generator}
\label{alg:random-td-net}
\begin{algorithmic}
    \STATE {\bfseries Input:} number of features $n_{p}$, discount factor $\gamma$, action set $\mathcal{A}$, depth $D$ and repeat $R$
    \STATE {\bfseries Output:} a network $G$
    \STATE $G \leftarrow$ an empty graph
    \STATE $roots \leftarrow$ an empty set
    \STATE $leaves \leftarrow$ an empty set
    \FOR{$i=1$ {\bfseries to} $n_{p}$}
        \STATE create a new feature node $f$ in $G$
        \STATE $roots \leftarrow roots \cup \{f\}$
        \STATE $leaves \leftarrow leaves \cup \{f\}$
        \STATE create a new prediction node $v$ in $G$
        \STATE $leaves \leftarrow leaves \cup \{v\}$
        \STATE add edge $<v, f, 1>$ to $G$
        \STATE add edge $<v, v, \gamma>$ to $G$
    \ENDFOR
    \FOR{$d=1$ {\bfseries to} $D$}
        \STATE $expanded \leftarrow$ an empty set
        \FOR{$a \in A$}
            \STATE $parent \leftarrow$ randomly select $R$ nodes from $leaves$ without replacement
            \FOR{$p \in parent$}
                \STATE create a new prediction node $v$ in $G$
                \STATE mark $v$ as conditioned on action $a$
                \STATE $expanded \leftarrow expanded \cup \{v\}$
                \STATE add edge $<v, p, 1>$ to $G$
                \STATE $f \leftarrow$ randomly select a node from $roots$
                \STATE add edge $<v, f, 1>$ to $G$
            \ENDFOR
        \ENDFOR
        \STATE $leaves \leftarrow expanded$
    \ENDFOR
\end{algorithmic}
\end{algorithm}

\clearpage
\section{Additional Empirical Results}

\begin{figure*}[h]
    \centering
    \subfloat[]{\includegraphics[width=\textwidth]{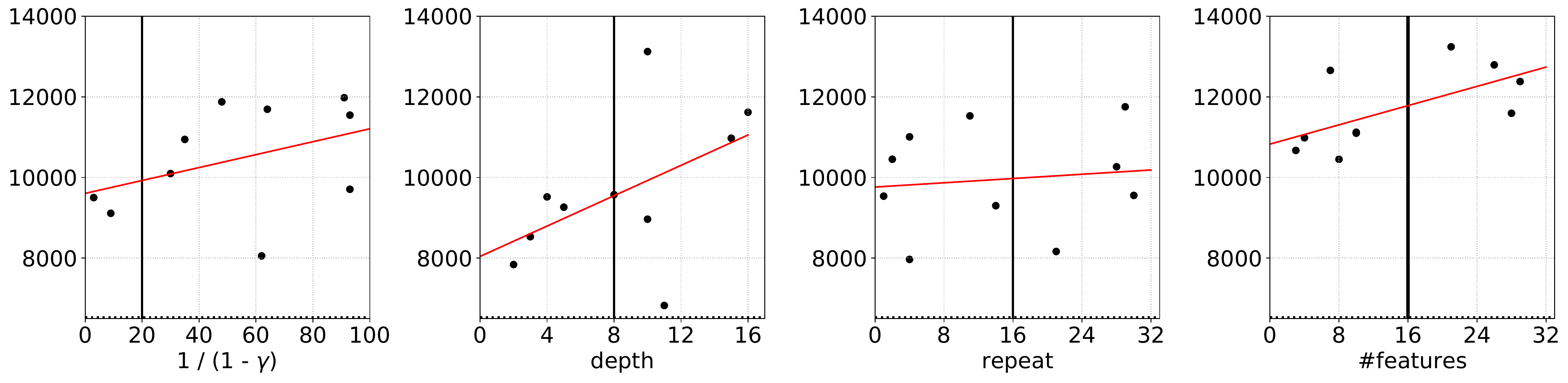}} \quad
    \subfloat[]{\includegraphics[width=\textwidth]{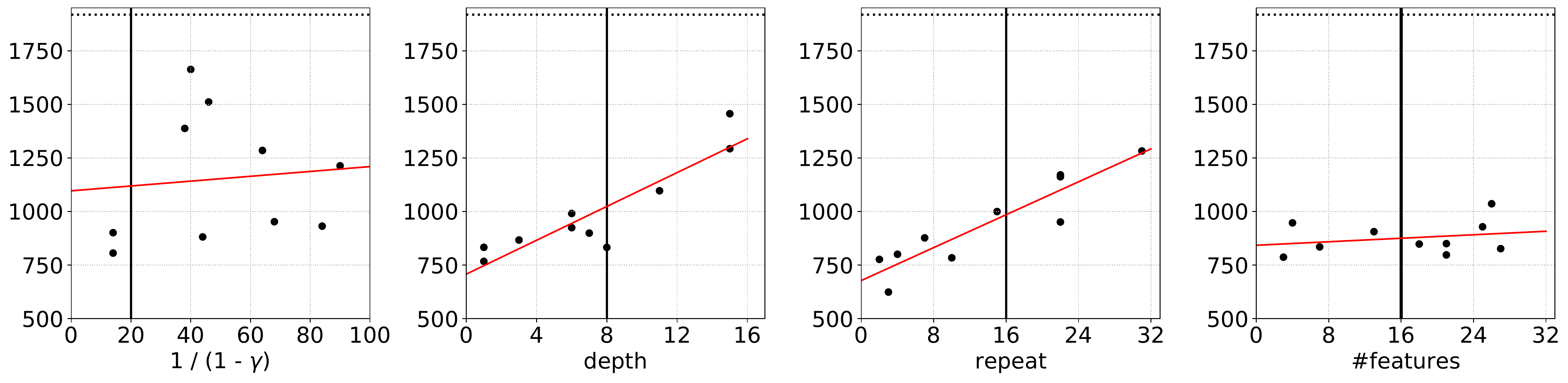}}
    \caption{Scatter plots of scores in \textbf{(a)} BeamRider and \textbf{(b)} SpaceInvaders obtained by {\name} with different hyperparameters. x-axis denotes the value of the hyperparameter. y-axis denotes the final game score after training for $200$ million frames. \revisedfinal{The red line in each panel is the line of best fit.} The dotted horizontal lines denote the performance of the end-to-end A2C baseline. The solid vertical lines denotes the values we used in our final experiments.}
    \label{fig:atari-robust}
\end{figure*}

\begin{figure*}[tb]
    \centering
    \includegraphics[height=0.96\textheight]{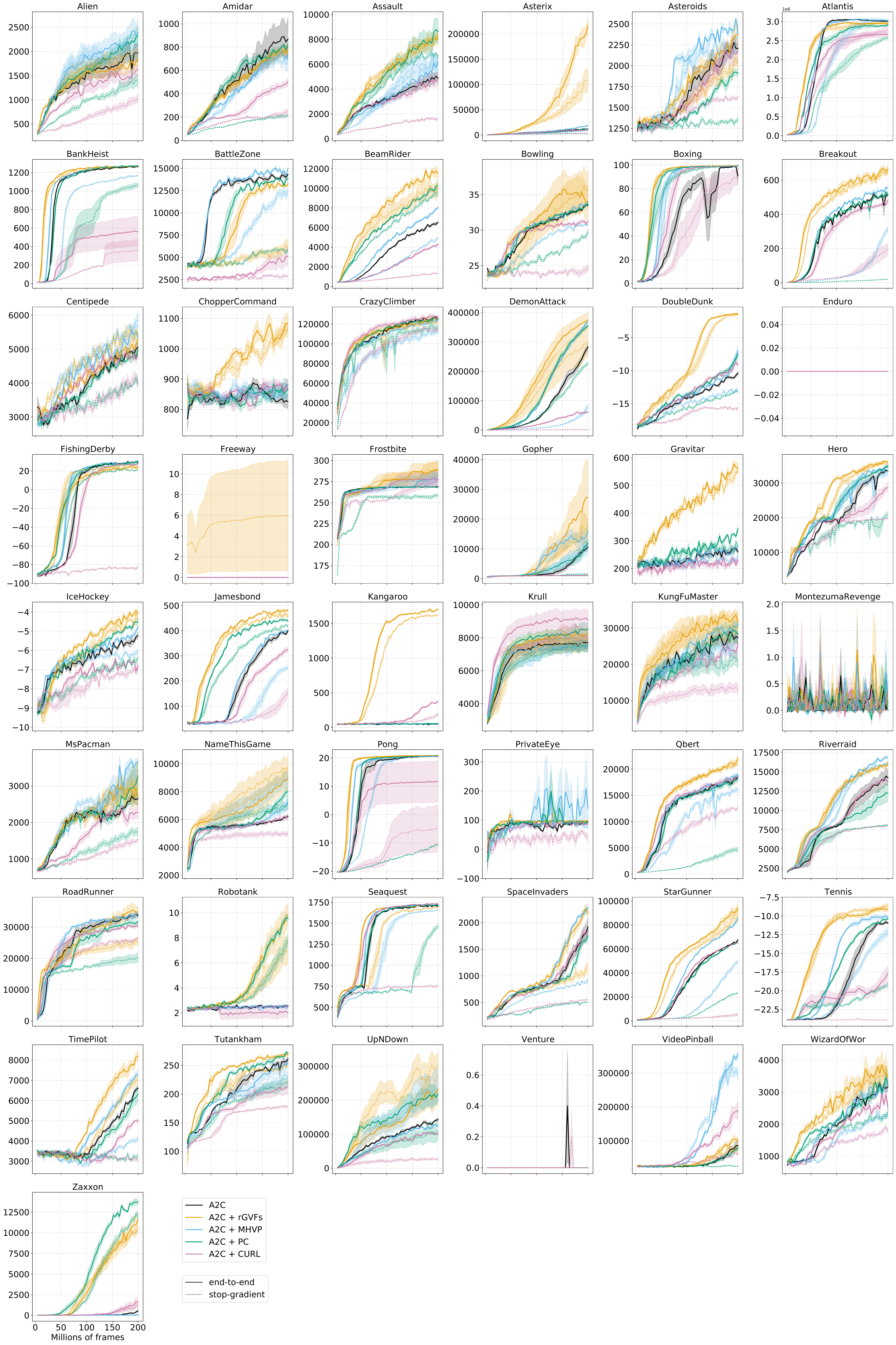}
    \caption{Learning curves in $49$ Atari games. The x-axis denotes the number of frames. Each dark curve is averaged over $5$ independent runs with different random seeds. The shaded area shows the standard error.}
    \label{fig:atari49-raw-score}
\end{figure*}

\begin{figure*}[h]
    \centering
    \includegraphics[width=0.96\textwidth]{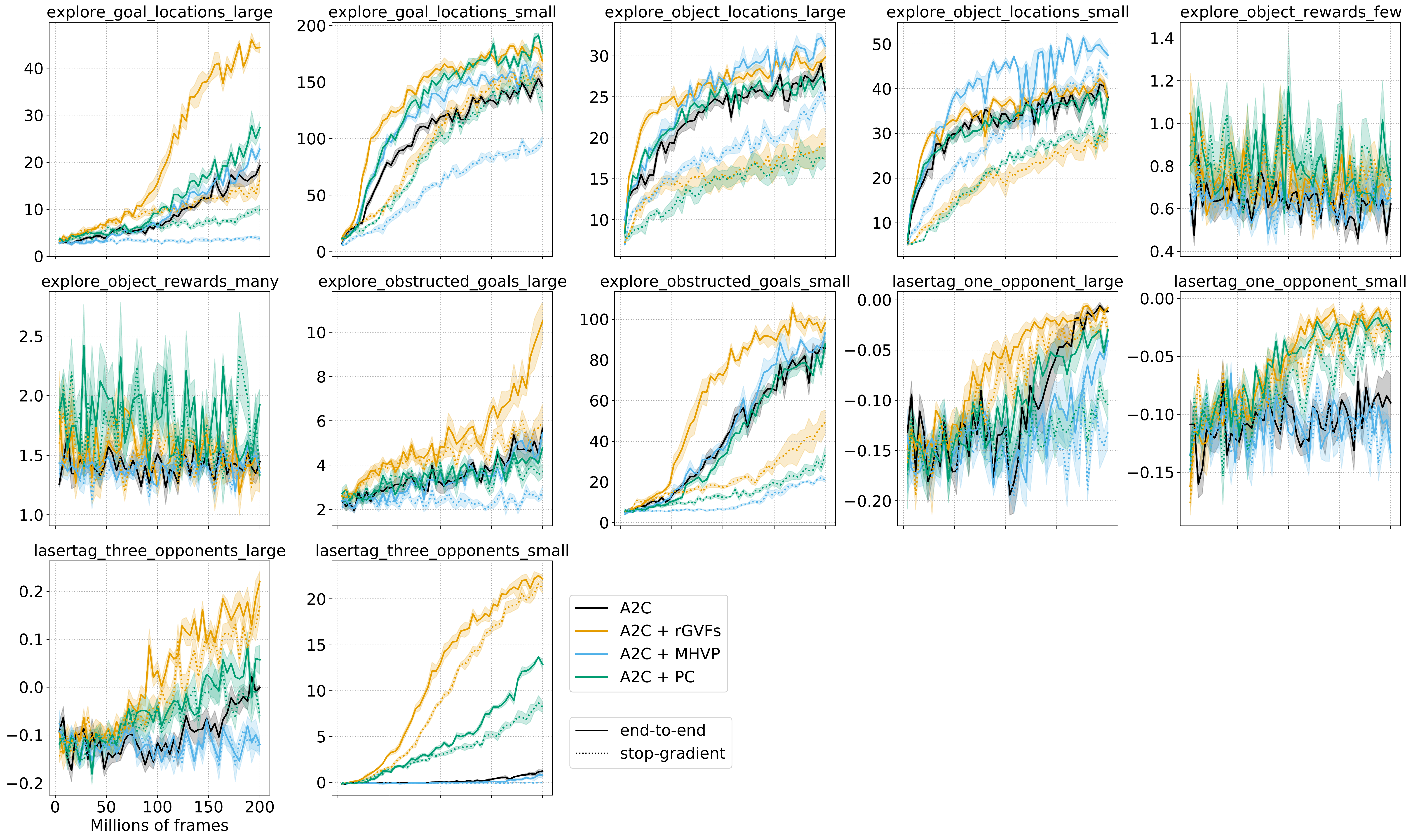}
    \caption{Learning curves in $12$ DeepMind Lab environments. The x-axis denotes the number of frames. Each dark curve is averaged over $5$ independent runs with different random seeds. The shaded area shows the standard error.}
    \label{fig:dmlab12-raw-score}
\end{figure*}


\end{document}